%% file: neurips_2023.tex
\documentclass{article}

\usepackage[square,sort,comma,numbers]{natbib}

\usepackage[final]{neurips_2023}

\usepackage[utf8]{inputenc} 
\usepackage[T1]{fontenc}    
\usepackage{url}            
\usepackage{booktabs}       
\usepackage{amsfonts}       
\usepackage{nicefrac}       
\usepackage{microtype}      
\usepackage{xcolor}         

\usepackage{booktabs}       
\usepackage{amsfonts}       
\usepackage{nicefrac}       
\usepackage{microtype}      
\usepackage{mathtools}
\usepackage{multirow}
\usepackage{bm}
\usepackage{epsfig}
\usepackage{graphicx}
\usepackage{caption}
\usepackage{subcaption}
\usepackage{amsthm}
\usepackage{amsmath}
\usepackage{amssymb}
\usepackage{enumitem}
\usepackage{makecell}
\usepackage{wrapfig}
\usepackage{indentfirst}
\usepackage{verbatim}
\usepackage{color}
\usepackage{setspace}
\usepackage{array}
\usepackage{booktabs}
\usepackage{stackengine}
\usepackage{algorithm}
\usepackage[algo2e,algoruled,boxed,lined]{algorithm2e}
\usepackage{algorithmicx}
\usepackage{graphicx}
\usepackage{xcolor}

\usepackage{caption}
\renewcommand{\captionlabelfont}{\scriptsize}

\usepackage[pagebackref=true,breaklinks=true,colorlinks=true,bookmarks=false]{hyperref}
\definecolor{deepred}{HTML}{940000}
\hypersetup{linkcolor=deepred}
\hypersetup{urlcolor  = [rgb]{0.4,0.15,0.95}}
\hypersetup{citecolor=[rgb]{0.4,0.15,0.95}}

\usepackage{colortbl}
\definecolor{Gray}{gray}{0.94}
\definecolor{Gray}{gray}{0.94}

\newlength\savewidth\newcommand\shline{\noalign{\global\savewidth\arrayrulewidth
  \global\arrayrulewidth 1pt}\hline\noalign{\global\arrayrulewidth\savewidth}}

\usepackage{pifont}
\usepackage{tocloft}
\usepackage[toc,page,header]{appendix}
\usepackage{adjustbox}
\usepackage{minitoc}

\renewcommand \thepart{}
\renewcommand \partname{}

\newcommand{\norm}[1]{\left\lVert#1\right\rVert}

\newcommand{\eg}{\emph{e.g.}}
\newcommand{\ie}{\emph{i.e.}}

\renewcommand{\captionlabelfont}{\scriptsize}

\newcommand\blfootnote[1]{%
  \begingroup
  \renewcommand\thefootnote{}\footnote{#1}%
  \addtocounter{footnote}{-1}%
  \endgroup
}

\title{\fontsize{14.5pt}{\baselineskip}\selectfont Controlling Text-to-Image Diffusion by Orthogonal Finetuning}

\author{%
  \vspace{-5mm}\\
  \textbf{\small Zeju Qiu\textsuperscript{1,*}~~~~Weiyang Liu\textsuperscript{1,2,*,\textdagger}~~~~Haiwen Feng\textsuperscript{1}~~~~Yuxuan Xue\textsuperscript{3}~~~~Yao Feng\textsuperscript{1}~~~~Zhen Liu\textsuperscript{1,4}}\\
  \textbf{\small Dan Zhang\textsuperscript{3,5}~~~~Adrian Weller\textsuperscript{2,6}~~~~Bernhard Sch\"olkopf\textsuperscript{1}}\\[0.5mm]
  \small \textsuperscript{1}MPI for Intelligent Systems - T\"ubingen~~~~\textsuperscript{2}University of Cambridge~~~~\textsuperscript{3}University of T\"ubingen\\
  \small \textsuperscript{4}Mila, Universit\'e de Montr\'eal~~~~\textsuperscript{5}Bosch Center for Artificial Intelligence~~~~\textsuperscript{6}The Alan Turing Institute\\
  \small \textsuperscript{*}Equal contribution~~~~~\textsuperscript{\textdagger}Project lead~~~~~\href{https://oft.wyliu.com}{\url{oft.wyliu.com}}
}

\begin{document}

\doparttoc
\faketableofcontents

\maketitle

\blfootnote{This work was finished when ZQ was a research intern hosted by WL at MPI for Intelligent Systems.}
\vspace{-8.5mm}
\begin{abstract}
\vspace{-0.5mm}
Large text-to-image diffusion models have impressive capabilities in generating photorealistic images from text prompts. How to effectively guide or control these powerful models to perform different downstream tasks becomes an important open problem. To tackle this challenge, we introduce a principled finetuning method -- Orthogonal Finetuning~(OFT), for adapting text-to-image diffusion models to downstream tasks. Unlike existing methods, OFT can provably preserve hyperspherical energy which characterizes the pairwise neuron relationship on the unit hypersphere. We find that this property is crucial for preserving the semantic generation ability of text-to-image diffusion models. To improve finetuning stability, we further propose Constrained Orthogonal Finetuning (COFT) which imposes an additional radius constraint to the hypersphere. Specifically, we consider two important finetuning text-to-image tasks: subject-driven generation where the goal is to generate subject-specific images given a few images of a subject and a text prompt, and controllable generation where the goal is to enable the model to take in additional control signals. We empirically show that our OFT framework outperforms existing methods in generation quality and convergence speed.

\end{abstract}

\vspace{-1.25mm}

\section{Introduction}
\vspace{-1mm}

Recent text-to-image diffusion models~\cite{saharia2022photorealistic,ramesh2022hierarchical,rombach2022high} achieve impressive performance in text-guided control for high-fidelity image generation. Despite strong results, text guidance can still be ambiguous and insufficient to provide fine-grained and accurate control to the generated images. To address this shortcoming, we target two types of text-to-image generation tasks in this paper:

\vspace{0.5mm}
\begin{itemize}[leftmargin=*,nosep]
\setlength\itemsep{0.4em}
    \item \textbf{Subject-driven generation}~\cite{ruiz2023dreambooth}: Given just a few images of a subject, the task is to generate images of the same subject in a different context using the guidance of a text prompt.
    \item \textbf{Controllable generation}~\cite{zhang2023adding,mou2023t2i}: Given an additional control signal (\eg, canny edges, segmentation maps), the task is to generate images following such a control signal and a text prompt.
\end{itemize}
\vspace{0.5mm}

Both tasks essentially boil down to how to effectively finetune text-to-image diffusion models without losing the pretraining generative performance. We summarize the desiderata for an effective finetuning method as: (1) \emph{training efficiency}: having fewer trainable parameters and number of training epochs, and (2) \emph{generalizability preservation}: preserving the high-fidelity and diverse generative performance. To this end, finetuning is typically done either by updating the neuron weights by a small learning rate (\eg, \cite{ruiz2023dreambooth}) or by adding a small component with re-parameterized neuron weights (\eg, \cite{hulora2022,zhang2023adding}). Despite simplicity, neither finetuning strategy is able to guarantee the preservation of pretraining generative performance. There is also a lack of principled understanding towards designing a good finetuning strategy and finding suitable hyperparameters such as the number of training epochs. A key difficulty is the lack of a measure for quantifying the preservation of pretrained generative ability. Existing finetuning methods implicitly assume that a smaller Euclidean distance between the finetuned model and the pretrained model indicates better preservation of the pretrained ability. Due to the same reason, finetuning methods typically work with a very small learning rate. While this assumption may occasionally hold, we argue that the Euclidean difference to the pretrained model alone is unable to fully capture the degree of semantic preservation, and therefore a more structural measure to characterize the difference between the finetuned model and the pretrained model can greatly benefit the preservation of pretraining performance as well as finetuning stability. 

Inspired by the empirical observation that hyperspherical similarity encodes semantic information well~\cite{liu2017deep,liu2018decoupled,chen2020angular}, we use hyperspherical energy~\cite{liu2018learning} to characterize the pairwise relational structure among neurons. Hyperspherical energy is defined as the sum of hyperspherical similarity (\eg, cosine similarity) between all pairwise neurons in the same layer, capturing the level of neuron uniformity on the unit hypersphere~\cite{liu2021learning}. We hypothesize that a good finetuned model should have a minimal difference in hyperspherical energy compared to the pretrained model. A naive way is to add a regularizer such that the hyperspherical energy remains the same during the finetuning stage, but there is no guarantee that the hyperspherical energy difference can be well minimized. Therefore, we take advantage of an invariance property of hyperspherical energy -- the pairwise hyperspherical similarity is provably preserved if we apply the same orthogonal transformation for all neurons. Motivated by such an invariance, we propose Orthogonal Finetuning~(OFT) which adapts large text-to-image diffusion models to a downstream task without changing its hyperspherical energy. The central idea is to learn a layer-shared orthogonal transformation for neurons such that their pairwise angles are preserved. OFT can also be viewed as adjusting the canonical coordinate system for the neurons in the same layer. By jointly taking into consideration that smaller Euclidean distance between the finetuned model and the pretrained model implies better preservation of pretraining performance, we further propose an OFT variant -- Constrained Orthogonal Finetuning~(COFT) which constrains the finetuned model within the hypersphere of a fixed radius centered on the pretrained neurons.

\begin{figure}[t]
    \centering
    \setlength{\abovecaptionskip}{4pt}
    \setlength{\belowcaptionskip}{-8.5pt}
    \renewcommand{\captionlabelfont}{\scriptsize}
    \vspace{-2.25pt}
    \includegraphics[width=\textwidth]{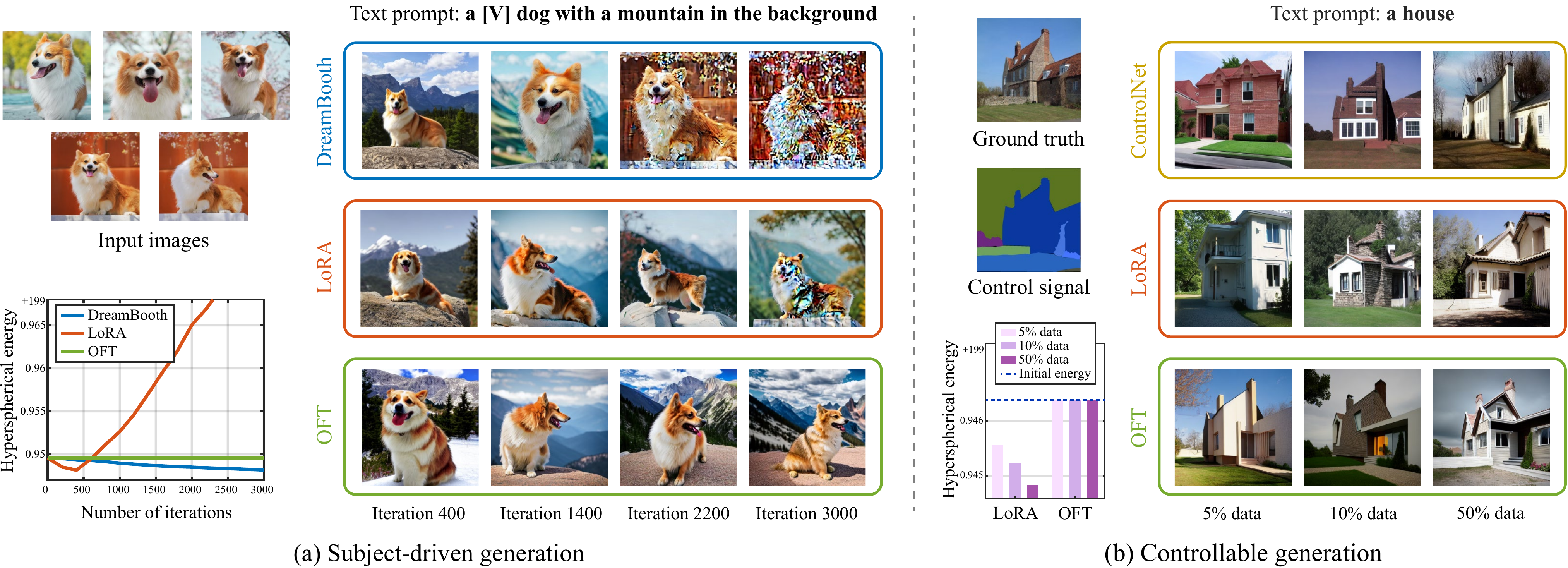}
    \caption{\scriptsize (a) Subject-driven generation: OFT preserves the hyperspherical energy and yields more stable finetuning performance across different number of iterations, while both DreamBooth~\cite{ruiz2023dreambooth} and LoRA~\cite{hulora2022} do not. OFT can preserve hyperspherical energy and perform stable finetuning, while both LoRA and DreamBooth are unable. (b) Controllable generation: OFT is more sample-efficient in training and converges well with only 5\% of the original dataset, while both ControlNet~\cite{zhang2023adding} and LoRA~\cite{hulora2022} cannot converge until 50\% of the data is present. The hyperspherical energy comparison between LoRA and OFT is fair, since they finetune the same layers. ControlNet uses a different layer finetuning strategy, so its hyperspherical energy is not comparable. The detailed settings are given in the experiment section and Appendix~\ref{app:settings}.}
    \label{fig:teaser}
\end{figure}

The intuition for why orthogonal transformation works for finetuning neurons partially comes from 2D Fourier transform, with which an image can be decomposed as magnitude and phase spectrum. The phase spectrum, which is angular information between input and basis, preserves the major part of semantics. For example, the phase spectrum of an image, along with a random magnitude spectrum, can still reconstruct the original image without losing its semantics. This phenomenon suggests that changing the neuron directions is the key to semantically modifying the generated image (which is the goal of both subject-driven and controllable generation). However, changing neuron directions with a large degree of freedom will inevitably destroy the pretraining generative performance. To constrain the degree of freedom, we propose to preserve the angle between any pair of neurons, largely based on the hypothesis that the angles between neurons are crucial for representing the knowledge of neural networks. With this intuition, it is natural to learn layer-shared orthogonal transformation for neurons in each layer such that the hyperspherical energy stays unchanged.

We also draw inspiration from orthogonal over-parameterized training~\cite{liu2021orthogonal} which trains classification neural networks from scratch by orthogonally transforming a randomly initialized neural network. This is because a randomly initialized neural network yields a provably small hyperspherical energy in expectation and the goal of \cite{liu2021orthogonal} is to keep hyperspherical energy small during training (small energy leads to better generalization in classification~\cite{liu2018learning,lin2020regularizing}). \cite{liu2021orthogonal} shows that orthogonal transformation is sufficiently flexible to train generalizable neural networks for classification problems. In contrast, we focus on finetuning text-to-image diffusion models for better controllability and stronger downstream generative performance. 
We emphasize the difference between OFT and \cite{liu2021orthogonal} in two aspects. First, while \cite{liu2021orthogonal} is designed to minimize the hyperspherical energy, OFT aims to preserve the same hyperspherical energy as the pretrained model so that the intrinsic pretrained structure will not be destroyed by finetuning. In the case of finetuning diffusion models, minimizing hyperspherical energy could destroy the original semantic structures. Second, OFT seeks to minimize the deviation from the pretrained model, which leads to the constrained variant. In contrast, \cite{liu2021orthogonal} imposes no such constraints. The key to finetuning is to find a good trade-off between flexibility and stability, and we argue that our OFT framework effectively achieves this goal. Our contributions are listed below:

\vspace{0.25mm}
\begin{itemize}[leftmargin=*,nosep]
\setlength\itemsep{0.4em}
    \item We propose a novel finetuning method -- Orthogonal Finetuning for guiding text-to-image diffusion models towards better controllability. To further improve stability, we propose a constrained variant which limits the angular deviation from the pretrained model.
    \item Compared to existing finetuning methods, OFT performs model finetuning while provably preserving the hyperspherical energy, which we empirically find to be an important measure of the generative semantic preservation of the pretrained model.
    \item We apply OFT to two tasks: subject-driven generation and controllable generation. We conduct a comprehensive empirical study and demonstrate significant improvement over prior work in terms of generation quality, convergence speed and finetuning stability. Moreover, OFT achieves better sample efficiency, as it converges well with a much smaller number of training images and epochs.
    \item For controllable generation, we introduce a new control consistency metric to evaluate the controllability. This core idea is to estimate the control signal from the generated image and then compare it with the origin control signal. The metric further validates the strong controllability of OFT.
\end{itemize}
\vspace{0.5mm}



\vspace{-2mm}
\section{Related Work}
\vspace{-1.46mm}

\textbf{Text-to-image diffusion models}. Tremendous progress~\cite{saharia2022photorealistic,ramesh2022hierarchical,rombach2022high,gu2022vector,nichol2022glide} has been made in text-to-image generation, largely thanks to the rapid development in diffusion-based generative models~\cite{ho2020denoising,song2021score,song2021denoising,dhariwal2021diffusion} and vision-language representation learning~\cite{su2020vlbert,lu2019vilbert,radford2021learning,wang2021simvlm,li2022blip,li2023blip,alayrac2022flamingo,schuhmann2022laion}. GLIDE~\cite{nichol2022glide} and Imagen~\cite{saharia2022photorealistic} train diffusion models in the pixel space. GLIDE trains the text encoder jointly with a diffusion prior using paired text-image data, while Imagen uses a frozen pretrained text encoder. Stable Diffusion~\cite{rombach2022high} and DALL-E2~\cite{ramesh2022hierarchical} train diffusion models in the latent space. Stable Diffusion uses VQ-GAN~\cite{esser2021taming} to learn a visual codebook as its latent space, while DALL-E2 adopts CLIP~\cite{radford2021learning} to construct a joint latent embedding space for representing images and text. Other than diffusion models, generative adversarial networks~\cite{reed2016generative,zhang2017stackgan,xu2018attngan,li2019controllable} and autoregressive models~\cite{ramesh2021zero,ding2021cogview,wu2022nuwa,yuscaling} have also been studied in text-to-image generation. OFT is inherently a model-agnostic finetuning approach and can be applied to any text-to-image diffusion model.

\textbf{Subject-driven generation}. To prevent subject modification, \cite{avrahami2022blended,nichol2022glide} consider a given mask from users as an additional condition. Inversion methods~\cite{choi2021ilvr,dhariwal2021diffusion,ramesh2022hierarchical,gal2022image} can be applied to modify the context without changing the subject. \cite{hertz2022prompt} can perform local and global editing without input masks. The methods above are unable to well preserve identity-related details of the subject. In Pivotal Tuning~\cite{roich2022pivotal}, a generator is finetuned around an initial inverted latent code with an additional regularization to preserve the identity. Similarly, \cite{nitzan2022mystyle} learns a personalized generative face prior from a collection of a person's face images. \cite{casanova2021instance} can generate difference variations of an instance, but it may lose the instance-specific details. With a customized token and a few subject images, DreamBooth~\cite{ruiz2023dreambooth} finetunes the text-to-image diffusion model using a reconstruction loss and a class-specific prior preservation loss. OFT adopts the DreamBooth framework, but instead of performing naive finetuning with a small learning rate, OFT finetunes the model with orthogonal transformations.

\textbf{Controllable generation}. The task of image-to-image translation can be viewed as a form of controllable generation, and previous methods mostly adopt conditional generative adversarial networks~\cite{isola2017image,zhu2017toward,wang2018high,park2019semantic,choi2018stargan}. Diffusion models are also used for image-to-image translation~\cite{saharia2022palette,voynov2022sketch,wang2022pretraining}. More recently, ControlNet~\cite{zhang2023adding} proposes to control a pretrained diffusion model by finetuning and adapting it to additional control signals and achieves impressive controllable generation performance. Another concurrent and similar work, T2I-Adapter~\cite{mou2023t2i}, also finetunes a pretrained diffusion model in order to gain stronger controllability for the generated images. Following the same task setting in \cite{zhang2023adding,mou2023t2i}, we apply OFT to finetune pretrained diffusion models, yielding consistently better controllability with fewer training data and less finetuning parameters. More significantly, OFT does not introduce any additional computational overhead during test-time inference.

\textbf{Model finetuning}. Finetuning large pretrained models on downstream tasks has been increasingly 
popular nowadays~\cite{devlin2018bert,brown2020language,he2022masked}. As a form of finetuning, adaptation methods (\eg, \cite{hulora2022,houlsby2019parameter,pfeiffer2020adapterfusion}) are heavily studied in natural language processing. LoRA~\cite{hulora2022} is the most relevant work to OFT, and it assumes a low-rank structure for the additive weight update during finetuning. In contrast, OFT uses layer-shared orthogonal transformation to update neuron weights in a multiplicative manner, and it provably preserves the pair-wise angles among neurons in the same layer, yielding better stability.

\vspace{-1.6mm}
\section{Orthogonal Finetuning}
\vspace{-1.15mm}

\subsection{Why Does Orthogonal Transformation Make Sense?}
\vspace{-.9mm}

We start by discussing why orthogonal transformation is desirable in finetuning text-to-image diffusion models. We decompose this question into two smaller ones: (1) why we want to finetune the angle of neurons (\ie, direction), and (2) why we adopt orthogonal transformation to finetune angles.

\begin{wrapfigure}{r}{0.395\linewidth}
\vspace{-1em}
\centering
\includegraphics[width=0.975\linewidth]{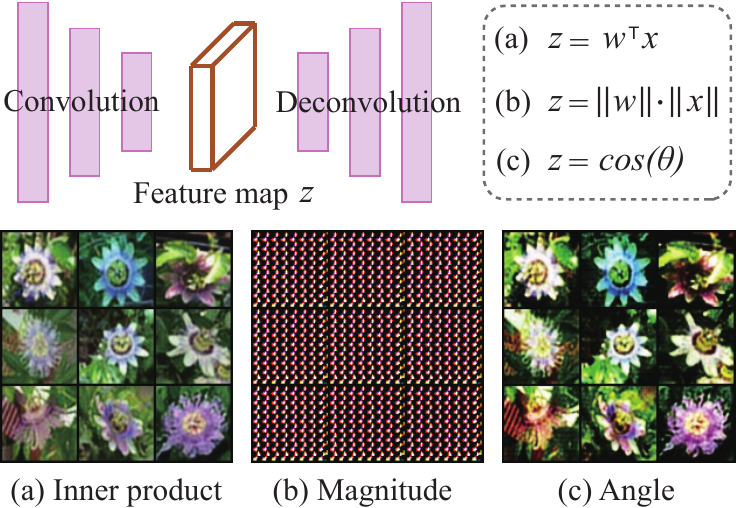}
  \vspace{-0.3em}
    \caption{\scriptsize A toy experiment to demonstrate the importance of angular information. The autoencoder is trained in a standard way using inner product activation, and (a) shows the standard reconstruction. In testing, the angular information of neurons alone can well recover the input image, even if the autoencoder is not trained with angles.}
    \label{fig:toy_ae}
\vspace{-0.5em}
\end{wrapfigure}

For the first question, we draw inspiration from the empirical observation in \cite{liu2018decoupled,chen2020angular} that angular feature difference well characterizes the semantic gap. SphereNet~\cite{liu2017deep} shows that training a neural network with all neurons normalized onto a unit hypersphere yields comparable capacity and even better generalizability, implying that the direction of neurons can fully capture the most important information from data. To better demonstrate the importance of neuron angles, we conduct a toy experiment in Figure~\ref{fig:toy_ae} where we train a standard convolutional autoencoder on some flower images. In the training stage, we use the standard inner product to produce the feature map ($z$ denotes the element output of the convolution kernel $\bm{w}$ and $\bm{x}$ is the input in the sliding window). In the testing stage, we compare three ways to generate the feature map: (a) the inner product used in training, (b) the magnitude information, and (c) the angular information. The results in Figure~\ref{fig:toy_ae} show that the angular information of neurons can almost perfectly recover the input images, while the magnitude of neurons contains no useful information. We emphasize that we do not apply the cosine activation (c) during training, and the training is done only based on inner product. The results imply that the angles (directions) of neurons play the major role in storing the semantic information of the input images. In order to modify the semantic information of images, finetuning the neuron directions will likely be more effective.

\begin{wrapfigure}{r}{0.3175\linewidth}
\vspace{-1.4em}
\centering
\includegraphics[width=0.98\linewidth]{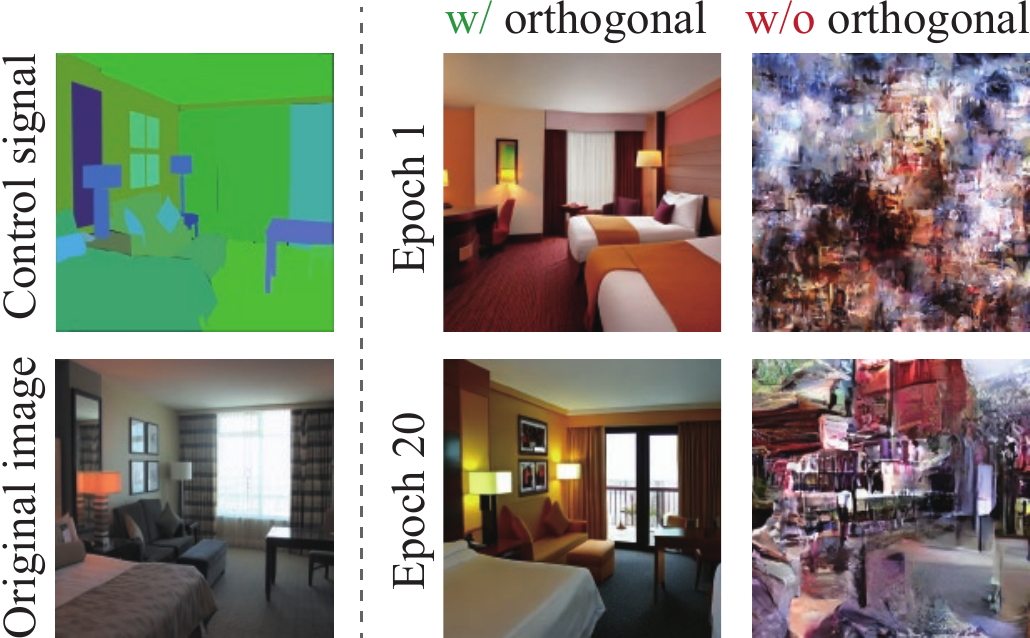}
  \vspace{-0.4em}
    \caption{\scriptsize Controllable generation with or without orthogonality. Middle column is from the original OFT, and the right column is given by OFT without the orthogonality constraint.}
    \label{fig:ortho}
\vspace{-0.6em}
\end{wrapfigure}

For the second question, the simplest way to finetune direction of neurons is to simultaneously rotate / reflect all the neurons (in the same layer), which naturally brings in orthogonal transformation. It may be more flexible to use some other angular transformation that rotates different neurons with different angles, but we find that orthogonal transformation is a sweet spot between flexibility and regularity. Moreover, \cite{liu2021orthogonal} shows that orthogonal transformation is sufficiently powerful for learning neural networks. To support our argument, we perform an experiment to demonstrate the effective regularization induced by the orthogonality constraint. We perform the controllable generation experiment using the setting of ControlNet~\cite{zhang2023adding}, and the results are given in Figure~\ref{fig:ortho}. We can observe that our standard OFT performs quite stably and achieves accurate control after the training is finished (epoch 20). In comparison, OFT without the orthogonality constraint fails to generate any realistic image and achieve no control effect. The experiment validates the importance of the orthogonality constraint in OFT.

\vspace{-1.35mm}
\subsection{General Framework}
\vspace{-1mm}

\begin{wrapfigure}{r}{0.565\linewidth}
\vspace{-1.2em}
\centering
\includegraphics[width=0.995\linewidth]{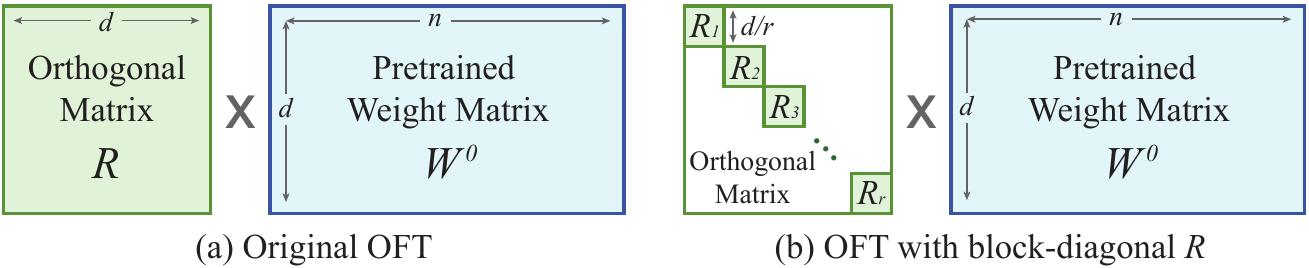}
  \vspace{-1.5em}
    \caption{\scriptsize (a) Original OFT without a diagonal structure. (b) OFT with $r$ diagonal blocks of the same size. When $r=1$, the case of (b) recovers the case of (a).}
    \label{fig:block}
\vspace{-0.35em}
\end{wrapfigure}

The conventional finetuning strategy typically uses gradient descent with a small learning rate to update a model (or certain layers of a model). The small learning rate implicitly encourages a small deviation from the pretrained model, and the standard finetuning essentially aims to train the model while implicitly minimizing $\thickmuskip=2mu \medmuskip=2mu \| \bm{M} - \bm{M}^0\|$ where $\bm{M}$ is the finetuned model weights and $\bm{M}^0$ is the pretrained model weights. This implicit constraint makes intuitive sense, but it can still be too flexible for finetuning a large model. To address this, LoRA introduces an additional low-rank constraint for the weight update, \ie, $\thickmuskip=2mu \medmuskip=2mu \text{rank}(\bm{M} - \bm{M}^0)=r'$ where $r'$ is set to be some small number. Different from LoRA, OFT introduces a constraint for the pair-wise neuron similarity: $\thickmuskip=2mu \medmuskip=2mu \| \text{HE}(\bm{M})-\text{HE}(\bm{M}^0) \|=0$ where $\text{HE}(\cdot)$ denotes hyperspherical energy of a weight matrix. As an illustrative example, we consider a fully connected layer $\thickmuskip=2mu \medmuskip=2mu \bm{W}=\{\bm{w}_1,\cdots,\bm{w}_n\}\in\mathbb{R}^{d\times n}$ where $\bm{w}_i\in\mathbb{R}^{d}$ is the $i$-th neuron ($\bm{W}^0$ is the pretrained weights). The output vector $\thickmuskip=2mu \medmuskip=2mu \bm{z}\in\mathbb{R}^n$ of this fully connected layer is computed by $\thickmuskip=2mu \medmuskip=2mu \bm{z}=\bm{W}^\top\bm{x}$ where $\thickmuskip=2mu \medmuskip=2mu \bm{x}\in\mathbb{R}^d$ is the input vector. OFT can be interpreted as minimizing the hyperspherical energy difference between the finetuned model and the pretrained model:
\begin{equation}\label{eq:oft_principle}
\footnotesize
\min_{\bm{W}} \left\| \text{HE}(\bm{W})-\text{HE}(\bm{W}^0)\right\|~~\Leftrightarrow~~\min_{\bm{W}} \bigg{\|} \sum_{i\neq j} \|\hat{\bm{w}}_i-\hat{\bm{w}}_j\|^{-1}-\sum_{i\neq j} \|\hat{\bm{w}}^0_i-\hat{\bm{w}}^0_j\|^{-1}\bigg{\|}
\end{equation}
where $\thickmuskip=2mu \medmuskip=2mu \hat{\bm{w}}_i:=\bm{w}_i/\|\bm{w}_i\|$ denotes the $i$-th normalized neuron, and the hyperspherical energy of a fully connected layer $\bm{W}$ is defined as $\thickmuskip=2mu \medmuskip=2mu \text{HE}(\bm{W}):= \sum_{i\neq j} \|\hat{\bm{w}}_i-\hat{\bm{w}}_j\|^{-1}$. One can easily observe that the attainable minimum is zero for Eq.~\eqref{eq:oft_principle}. The minimum can be achieved as long as $\bm{W}$ and $\bm{W}^0$ differ only up to a rotation or reflection, \ie, $\thickmuskip=2mu \medmuskip=2mu \bm{W}=\bm{R}\bm{W}^0$ in which $\thickmuskip=2mu \medmuskip=2mu \bm{R}\in\mathbb{R}^{d\times d}$ is an orthogonal matrix (The determinant $1$ or $-1$ means rotation or reflection, respectively). This is exactly the idea of OFT, that we only need to finetune the neural network by learning layer-shared orthogonal matrices to transform neurons in each layer. Formally, OFT seeks to optimize the orthogonal matrix $\thickmuskip=2mu \medmuskip=2mu \bm{R}\in\mathbb{R}^{d \times d}$ for a pretrained fully connected layer $\thickmuskip=2mu \medmuskip=2mu\bm{W}^0\in\mathbb{R}^{d\times n}$, changing the forward pass from $\thickmuskip=2mu \medmuskip=2mu\bm{z}=(\bm{W}^0)^\top\bm{x}$ to
\begin{equation}\label{eq:oft_general}
    \footnotesize
    \bm{z}=\bm{W}^\top\bm{x}=(\bm{R}\cdot\bm{W}^0)^\top\bm{x},~~~\text{s.t.}~\bm{R}^\top\bm{R}=\bm{R}\bm{R}^\top=\bm{I}
\end{equation}
where $\bm{W}$ denotes the OFT-finetuned weight matrix and $\bm{I}$ is an identity matrix. OFT is illustrated in Figure~\ref{fig:block}. Similar to the zero initialization in LoRA, we need to ensure OFT to finetune the pretrained model exactly from $\bm{W}^0$. To achieve this, we initialize the orthogonal matrix $\bm{R}$ to be an identity matrix so that the finetuned model starts with the pretrained weights. To guarantee the orthogonality of the matrix $\bm{R}$, we can use differential orthogonalization strategies discussed in \cite{liu2021orthogonal,lezcano2019cheap}. We will discuss how to guarantee the orthogonality in a computationally efficient way.

\vspace{-0.9mm}
\subsection{Efficient Orthogonal Parameterization}\label{sect:eff_ortho}
\vspace{-0.9mm}

Standard orthogonalization such as Gram-Schmidt method, despite differentiable, is often too expensive to compute in practice~\cite{liu2021orthogonal}. For better efficiency, we adopt Cayley parameterization to generate the orthogonal matrix. Specifically, we construct the orthogonal matrix with $\thickmuskip=2mu \medmuskip=2mu \bm{R}=(\bm{I}+\bm{Q})(I-\bm{Q})^{-1}$ where $\bm{Q}$ is a skew-symmetric matrix satisfying $\thickmuskip=2mu \medmuskip=2mu \bm{Q}=-\bm{Q}^\top$. Such an efficiency comes at a small price -- the Cayley parameterization can only produce orthogonal matrices with determinant $1$ which belongs to the special orthogonal group. Fortunately, we find that such a limitation does not affect the performance in practice. Even if we use Cayley transform to parameterize the orthogonal matrix, $\bm{R}$ can still be very parameter-inefficient with a large $d$. To address this, we propose to represent $\bm{R}$ with a block-diagonal matrix with $r$ blocks, leading to the following form:
\begin{equation}
    \footnotesize
    \bm{R}=\text{diag}(\bm{R}_1,\bm{R}_2,\cdots,\bm{R}_r)=\begin{bmatrix}
    \bm{R}_1\in \text{O}(\frac{d}{r}) &  &\\
    &\ddots & \\
    & & \bm{R}_r\in \text{O}(\frac{d}{r})
    \end{bmatrix}\in \text{O}(d)
\end{equation}
where $\text{O}(d)$ denotes the orthogonal group in dimension $d$, and $\thickmuskip=2mu \medmuskip=2mu \bm{R}\in\mathbb{R}^{d\times d}$ and $\thickmuskip=2mu \medmuskip=2mu \bm{R}_i\in\mathbb{R}^{d/r\times d/r},\forall i$ are orthogonal matrices. When $\thickmuskip=2mu \medmuskip=2mu r=1$, then the block-diagonal orthogonal matrix becomes a standard unconstrained one. For an orthogonal matrix with size $\thickmuskip=2mu \medmuskip=2mu d\times d$, the number of parameters is $\thickmuskip=2mu \medmuskip=2mu d(d-1)/2$, resulting in a complexity of $\mathcal{O}(d^2)$. For an $r$-block diagonal orthogonal matrix, the number of parameter is $\thickmuskip=2mu \medmuskip=2mu d(d/r-1)/2$, resulting in a complexity of $\thickmuskip=2mu \medmuskip=2mu \mathcal{O}(d^2/r)$. We can optionally share the block matrix to further reduce the number of parameters, \ie, $\thickmuskip=2mu \medmuskip=2mu\bm{R}_i=\bm{R}_j,\forall i\neq j$. This reduces the parameter complexity to $\mathcal{O}(d^2/r^2)$.  Despite all these strategies to improve parameter efficiency, we note that the resulting matrix $\bm{R}$ remains orthogonal, so there is no sacrifice in preserving hyperspherical energy.

We discuss how OFT compares to LoRA in terms of parameter efficiency. For LoRA with a low-rank parameter $r'$, we have its number of trainable parameters as $\thickmuskip=2mu \medmuskip=2mu r'(d+n)$. If we consider both $r$ and $r'$ to be dependent on the neuron dimension $d$ (\eg, $\thickmuskip=2mu \medmuskip=2mu r=r'=\alpha d$ where $\thickmuskip=2mu \medmuskip=2mu 0<\alpha\leq 1$ is some constant), then the parameter complexity of LoRA becomes $\mathcal{O}(d^2+dn)$ and the parameter complexity of OFT becomes $\mathcal{O}(d)$. We illustrate the difference in complexity between OFT and LoRA with a concrete example. Suppose we have a weight matrix with size $\thickmuskip=2mu \medmuskip=2mu 128\times 128$, LoRA has $2,048$ trainable parameters with $\thickmuskip=2mu \medmuskip=2mu r'=8$, while OFT has $960$ trainable parameters with $\thickmuskip=2mu \medmuskip=2mu r=8$ (no block sharing is applied).

\vspace{-1.5mm}
\subsection{Constrained Orthogonal Finetuning}
\vspace{-0.9mm}

We can further limit the flexibility of original OFT by constraining the finetuned model to be within a small neighborhood of the pretrained model. Specifically, COFT uses the following forward pass:
\begin{equation}\label{eq:coft_general}
    \footnotesize
    \bm{z}=\bm{W}^\top\bm{x}=(\bm{R}\cdot\bm{W}^0)^\top\bm{x},~~~\text{s.t.}~\bm{R}^\top\bm{R}=\bm{R}\bm{R}^\top=\bm{I},~\norm{\bm{R}-\bm{I}}\leq \epsilon
\end{equation}
which has an orthogonality constraint and an $\epsilon$-deviation constraint to an identity matrix. The orthogonality constraint can be achieved with the Cayley parameterization introduced in Section~\ref{sect:eff_ortho}. However, it is nontrivial to incorporate the $\epsilon$-deviation constraint to the Cayley-parameterized orthogonal matrix. To gain more insights on the Cayley transform, we apply the Neumann series to approximate $\thickmuskip=2mu \medmuskip=2mu \bm{R}=(\bm{I}+\bm{Q})(I-\bm{Q})^{-1}$ as $\thickmuskip=2mu \medmuskip=2mu \bm{R}\approx\bm{I}+2\bm{Q}+\mathcal{O}(\bm{Q}^2)$ (under the assumption that the Neumann \looseness=-1 {\parfillskip=0pt\par}

\begin{wrapfigure}{r}{0.635\linewidth}
\vspace{-0.65em}
\centering
\includegraphics[width=0.995\linewidth]{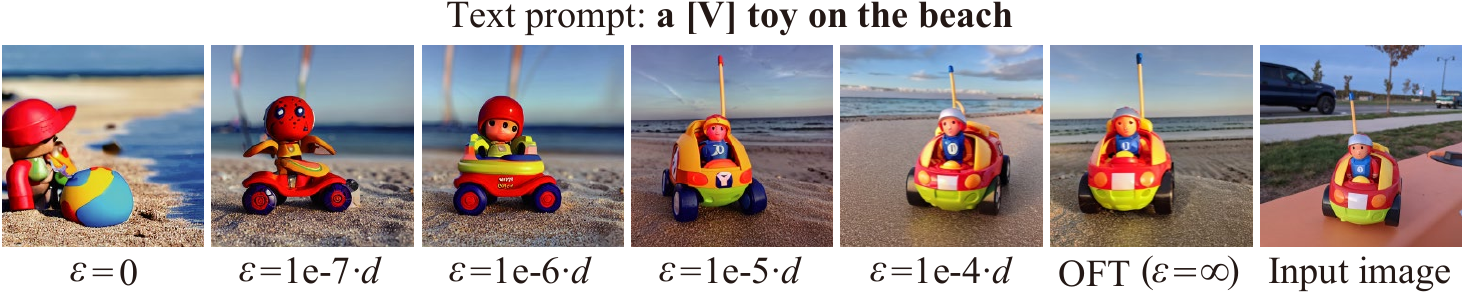}
  \vspace{-1.7em}
    \caption{\scriptsize How $\epsilon$ affects the flexibility of COFT in subject-driven generation.}
    \label{fig:eps_coft}
\vspace{-0.6em}
\end{wrapfigure}

\vspace{-1.625mm}

series converges in the operator norm). Therefore, we can move the constraint $\thickmuskip=2mu \medmuskip=2mu\|\bm{R}-\bm{I}\|\leq\epsilon$ inside the Cayley transform, and the equivalent constraint is $\thickmuskip=2mu \medmuskip=2mu \|\bm{Q}-\bm{0}\|\leq\epsilon'$ where $\bm{0}$ denotes an all-zero matrix and $\epsilon'$ is another error hyperparameter (different than $\epsilon$). The new constraint on the matrix $\bm{Q}$ can be easily enforced by projected gradient descent. To achieve identity initialization for the orthogonal matrix $\bm{R}$, we initialize $\bm{Q}$ as an all-zero matrix. COFT can be viewed as a combination of two explicit constraints: minimal hyperspherical energy difference and constrained deviation from the pretrained model. The second constraint is usually implicitly used by existing finetuning methods, but COFT makes it an explicit one. Despite the excellent performance of OFT, we observe that COFT yields even better finetuning stability than OFT due to this explicit deviation constraint. Figure~\ref{fig:eps_coft} provides an example on how $\epsilon$ affects the performance of COFT. We can observe that $\epsilon$ controls the flexibility of finetuning. With larger $\epsilon$, the COFT-finetuned model resembles the OFT-finetuned model. With smaller $\epsilon$, the COFT-finetuned model behaves increasingly similar to the pretrained text-to-image diffusion model.

\vspace{-1.5mm}
\subsection{Re-scaled Orthogonal Finetuning}
\vspace{-0.9mm}
We propose a simple extension to the original OFT by additionally learning a magnitude scaling coefficient for each neuron. This is motivated by the fact that re-scaling neurons does not change the hyperspherical energy (the magnitude will be normalized out). Specifically, we use the forward pass: $\thickmuskip=2mu \medmuskip=2mu\bm{z}=(\bm{R}\bm{W}^0\bm{D})^\top\bm{x}$\footnote{\textbf{Errata}: In the NeurIPS camera ready version, the forward pass of re-scaled OFT is mistakenly written as $\thickmuskip=2mu \medmuskip=2mu\bm{z}=(\bm{D}\bm{R}\bm{W}^0)^\top\bm{x}$. The original implementation is correct, so the results in Appendix~\ref{app:rescaled} are unaffected.} where $\thickmuskip=2mu \medmuskip=2mu \bm{D}=\text{diag}(s_1,\cdots,s_n)\in\mathbb{R}^{n\times n}$ is a learnable diagonal matrix with all the diagonal element $s_1,\cdots,s_n$ larger than zero. In contrast to OFT's original forward pass in Eq.~\eqref{eq:oft_general} where only $\bm{R}$ is learnable, we have both the diagonal matrix $\bm{D}$ and the orthogonal matrix $\bm{R}$ learnable. The re-scaled OFT further improves the flexibility of OFT with a neglectable number of additional parameters. We stick to the original OFT in the experiment to show the effectiveness of orthogonal transformation alone, but we find that the re-scaled OFT is generally better (see Appendix~\ref{app:rescaled}).

\vspace{-1.75mm}
\section{Intriguing Insights and Discussions}
\vspace{-1.5mm}

\textbf{OFT is agnostic to different architectures}. We can apply OFT to any type of neural network in principle. For Transformers, LoRA is typically applied to the attention weights~\cite{hulora2022}. To compare fairly to LoRA, we only apply OFT to finetune the attention weights in our experiments. Besides fully connected layers, OFT is also well suited for finetuning convolution layers, because the block-diagonal structure of $\bm{R}$ has interesting interpretations in convolution layers (unlike LoRA). When we use the same number of blocks as the number of input channels, each block only transforms a unique neuron channel, similar to learning depth-wise convolution kernels~\cite{chollet2017xception}. When all the blocks in $\bm{R}$ are shared, OFT transforms the neurons with an orthogonal matrix shared across channels. We conduct a preliminary study on finetuning convolution layers with OFT in Appendix~\ref{app:convolution}

\vspace{-0.4mm}

\textbf{Connection to LoRA}. By adding a low-rank matrix, LoRA prevents the information in the pretrained weight matrix from shifting dramatically. In contrast, OFT controls the transform that applies to the pretrained weight matrix to be orthogonal (full-rank), which prevents the transform to destroy the pretraining information. We can rewrite OFT's forward pass as $\thickmuskip=2mu \medmuskip=2mu \bm{z}=(\bm{R}\bm{W}^0)^\top\bm{x}=(\bm{W}^0+(\bm{R}-\bm{I})\bm{W}^0)^\top\bm{x}$ where $\thickmuskip=2mu \medmuskip=2mu (\bm{R}-\bm{I})\bm{W}^0$ is analogous to LoRA's low-rank weight update. Since $\bm{W}^0$ is typically full-rank, OFT also performs low-rank weight update when $\thickmuskip=2mu \medmuskip=2mu \bm{R}-\bm{I}$ is low-rank. Similar to LoRA that has a rank parameter $r'$, OFT has a diagonal block parameter $r$ to reduce the number of trainable parameters. More interestingly, LoRA and OFT represent two distinct ways to be parameter-efficient. LoRA exploits the low-rank structure to reduce the number of trainable parameters, while OFT takes a different route by exploiting the sparsity structure (\ie, block-diagonal orthogonality).

\vspace{-0.4mm}

\textbf{Why OFT converges faster}. On one hand, we can see from Figure~\ref{fig:toy_ae} that the most effective update to modify the semantics is to change neuron directions, which is exactly what OFT is designed for. On the other hand, OFT can be viewed as finetuning neurons on a smooth hypersphere manifold, which yields better optimization landscape. This is also empirically verified in \cite{liu2021orthogonal}.

\vspace{-0.4mm}

\textbf{Why not minimize hyperspherical energy}. A key difference to \cite{liu2021orthogonal} is that we do not aim to minimize hyperspherical energy. In classification problems, neurons without redundancy are desired. The minimum hyperspherical energy means all neurons are uniformly spaced around the hypersphere. This is not a meaningful objective for finetuning, as it may destroy the pretraining information.

\vspace{-0.4mm}

\textbf{Trade-off between flexibility and regularity in finetuning}. We discover an underlying trade-off between flexibility and regularity. Standard finetuning is the most flexible method, but it yields poor stability and easily causes model collapse. Being surprisingly simple, OFT finds a good balance between flexibility and regularity by preserving the pairwise neuron angles. The block-diagonal parameterization can also be viewed as a stronger regularization of the orthogonal matrix. 

\vspace{-0.4mm}

\textbf{No additional inference overhead}. Unlike ControlNet, our OFT framework introduces no additional inference overhead to the finetuned model. In the inference stage, we can simply multiply the learned orthogonal matrix $\bm{R}$ into the pretrained weight matrix $\bm{W}^0$ and obtain an equivalent weight matrix $\thickmuskip=2mu \medmuskip=2mu \bm{W}=\bm{R}\bm{W}^0$. Thus the inference speed is the same as the pretrained model.

\vspace{-2.05mm}
\section{Experiments and Results}
\vspace{-1.75mm}

\textbf{General settings}. In the experiment, we use Stable Diffusion v1.5~\cite{rombach2022high} as the pretrained text-to-image model. 
For fairness, we randomly pick generated images from each method. For subject-driven generation, we generally follow DreamBooth~\cite{ruiz2023dreambooth}. For controllable generation, we generally follow  ControlNet~\cite{zhang2023adding} and T2I-Adapter~\cite{mou2023t2i}. To ensure a fair comparison to LoRA, we only apply OFT or COFT to the same layer where LoRA is used. More results and details are given in Appendix~\ref{app:settings}.

\vspace{-1.4mm}
\subsection{Subject-driven Generation}
\vspace{-1.05mm}

\begin{wrapfigure}{r}{0.4\linewidth}
\vspace{-3.1em}
\centering
\includegraphics[width=0.995\linewidth]{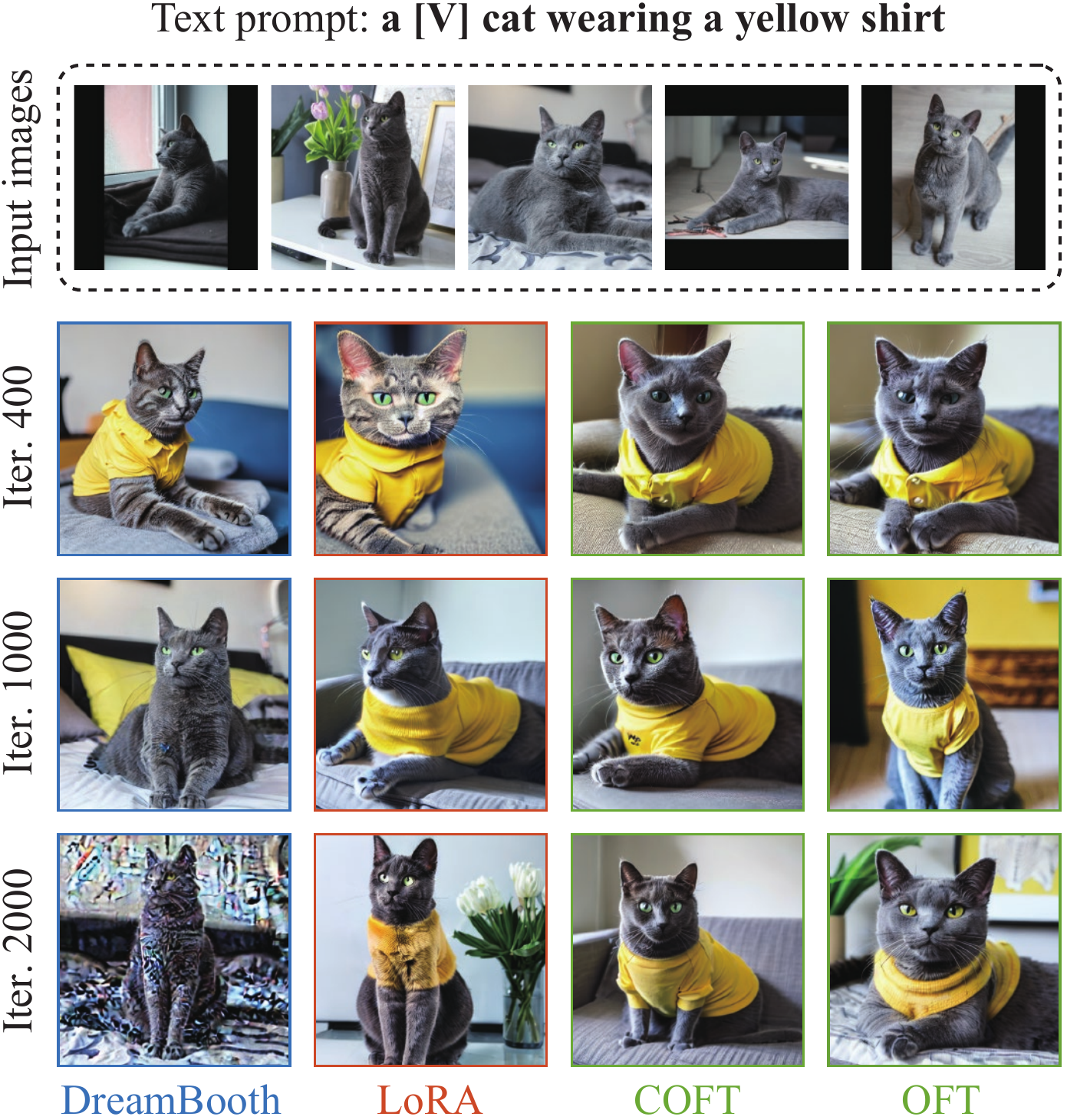}
  \vspace{-1.55em}
    \caption{\scriptsize Generated images across different iterations.}
    \label{fig:db_convergence}
\vspace{-1.1em}
\end{wrapfigure}

\textbf{Settings}. We use DreamBooth~\cite{ruiz2023dreambooth} and LoRA~\cite{hulora2022} as the baselines. All the methods adopt the same loss function as in DreamBooth. For DreamBooth and LoRA, we generally follow the original paper and use the best hyperparameter setup. More results are provided in Appendix~\ref{app:settings},\ref{app:coft_oft},\ref{app:more_qualitative},\ref{app:failure}.

\textbf{Finetuning stability and convergence}. We first evaluate the finetuning stability and the convergence speed for DreamBooth, LoRA, OFT and COFT. Results are given in Figure~\ref{fig:teaser} and Figure~\ref{fig:db_convergence}. We can observe that both COFT and OFT are able to finetune the diffusion model quite stably. After 400 iterations, both DreamBooth and OFT variants achieve good control, while LoRA fails to preserve the subject identity. After 2000 iterations, DreamBooth starts to generate collapsed images, and LoRA fails to generate yellow shirt (and instead generates yellow fur). In contrast, both OFT and COFT are still able to achieve stable and consistent control over the generated image. These results validate the fast yet stable convergence of our OFT framework in subject-driven generation. We note that the insensitivity to the number of finetuning iteration is quite important, since it can effectively alleviate the trouble of tuning the iteration number for different subjects. For both OFT and COFT, we can directly set a relatively large iteration number without carefully tuning it. For COFT with a proper $\epsilon$, both the learning rate and the iteration number become effortless to set.

\setlength{\columnsep}{11pt}
\begin{wraptable}{r}[0cm]{0pt}
    \scriptsize
    \centering
        \hspace{-2.4mm}
    \setlength{\tabcolsep}{3pt}
    \renewcommand{\arraystretch}{1.22}
    \renewcommand{\captionlabelfont}{\footnotesize}
        \begin{tabular}{lcccc}
        \specialrule{0em}{0pt}{-10.85pt}
        Method & DINO $\uparrow$ & CLIP-I $\uparrow$ & CLIP-T $\uparrow$ & LPIPS $\uparrow$\\
        \shline
        Real Images & 0.703 & 0.864 & - & 0.695 \\
        DreamBooth & 0.614 & 0.778 & \textbf{0.239} & 0.737 \\
        LoRA & 0.613 & 0.765 & 0.237 & 0.744 \\
        \rowcolor{Gray} COFT & 0.630 & 0.783 & 0.235 & 0.744 \\\rowcolor{Gray}
        OFT & \textbf{0.632} & \textbf{0.785} & 0.237 & \textbf{0.746}\\
        \specialrule{0em}{-3.5pt}{0pt}
        \end{tabular}
    \caption{\scriptsize Quantitative comparison of subject fidelity (DINO, CLIP-I), prompt fidelity (CLIP-T) and diversity metric (LPIPS). The evaluation images and prompts are the same as \cite{ruiz2023dreambooth} (25 subjects with 30 text prompts each subject).} \label{table:dreambooth_final}
\vspace{-2.9mm}
\end{wraptable}

\begin{figure}[t]
    \centering
    \setlength{\abovecaptionskip}{4pt}
    \setlength{\belowcaptionskip}{-8pt}
    \renewcommand{\captionlabelfont}{\scriptsize}
    \vspace{-1pt}
    \includegraphics[width=\textwidth]{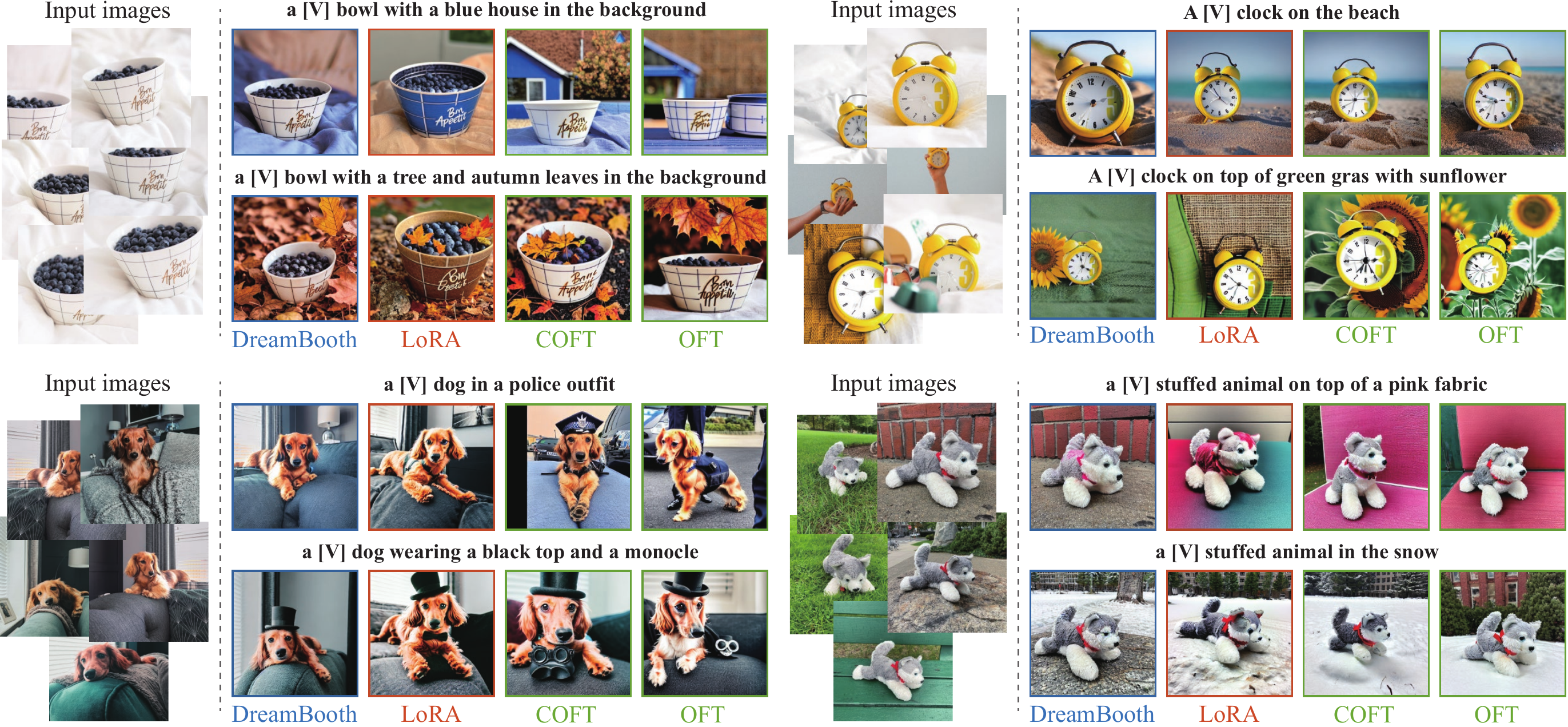}
    \caption{\scriptsize Qualitative comparison of subject-driven generation among DreamBooth, LoRA, COFT and OFT. Results are generated with the same finetuned model from each method. All examples are randomly picked. The figure is best viewed digitally, in color and significantly zoomed in.\looseness=-1}
    \label{fig:dreambooth_final}
\end{figure}

\textbf{Quantitative comparison}. Following \cite{ruiz2023dreambooth}, we conduct a quantitative comparison to evaluate subject fidelity (DINO~\cite{caron2021emerging}, CLIP-I~\cite{radford2021learning}), text prompt fidelity (CLIP-T~\cite{radford2021learning}) and sample diversity (LPIPS~\cite{zhang2018unreasonable}). CLIP-I computes the average pairwise cosine similarity of CLIP embeddings between generated and real images. DINO is similar to CLIP-I, except that we use ViT S/16 DINO embeddings. CLIP-T is the average cosine similarity of CLIP embeddings between text prompt and generated images. We also evaluate average LPIPS cosine similarity between generated images of the same subject with the same text prompt. Table~\ref{table:dreambooth_final} show that both COFT and OFT outperforms DreamBooth and LoRA in the DINO and CLIP-I metrics by a considerable margin, while achieving slightly better or comparable performance in prompt fidelity and diversity metric. For each method, we repeatedly finetune the same pretrained model with 30 different random seeds to minimize randomness. The results show that our OFT framework not only achieves better convergence and stability, but also yields consistently better final performance.

\textbf{Qualitative comparison}. To have a more intuitive understanding of OFT's benefits, we show some randomly picked examples for subject-driven generation in Figure~\ref{fig:dreambooth_final}. For a fair comparison, all the examples are generated from the same finetuned model using each method, so no text prompt will be separately optimized for its final results. For each method, we select the model that achieves the best validation CLIP metrics. From the results in Figure~\ref{fig:dreambooth_final}, we can observe that both OFT and COFT deliver excellent semantic subject preservation, while LoRA often fails to preserve the subject identity (\eg, LoRA completely loses the subject identity in the bowl example). In the meantime, both OFT and COFT have much more accurate control using text prompts, while DreamBooth, despite its preservation of subject identity, often fails to generate the image following the text prompt (\eg, the first row of the bowl example). The qualitative comparison demonstrates that our OFT framework achieves better controllability and subject preservation at the same time. Moreover, the number of iterations is not sensitive in OFT, so OFT performs well even with a large number of iterations, while neither DreamBooth nor LoRA can. More qualitative examples are given in Appendix~\ref{app:more_qualitative}. Moreover, we conduct a human evaluation in Appendix~\ref{app:human} which further validates the superiority of OFT.

\vspace{-1.5mm}
\subsection{Controllable Generation}
\vspace{-1.05mm}

\textbf{Settings}. We use ControlNet~\cite{zhang2023adding}, T2I-Adapter~\cite{mou2023t2i} and LoRA~\cite{hulora2022} as the baselines. We consider three challenging controllable generation tasks in the main paper: Canny edge to image~(C2I) on the COCO dataset~\cite{lin2014microsoft}, segmentation map to image~(S2I) on the ADE20K dataset~\cite{zhou2017scene} and landmark to face~(L2F) on the CelebA-HQ dataset~\cite{karras2018progressive,xia2021tedigan}. All the methods are used to finetune Stable Diffusion~(SD) v1.5 on these three datasets for 20 epochs. More results are given in Appendix~\ref{app:more_qualitative},\ref{app:more_control_task},\ref{app:failure}.

\begin{wrapfigure}{r}{0.26\linewidth}
\vspace{-1.25em}
\centering
\includegraphics[width=0.999\linewidth]{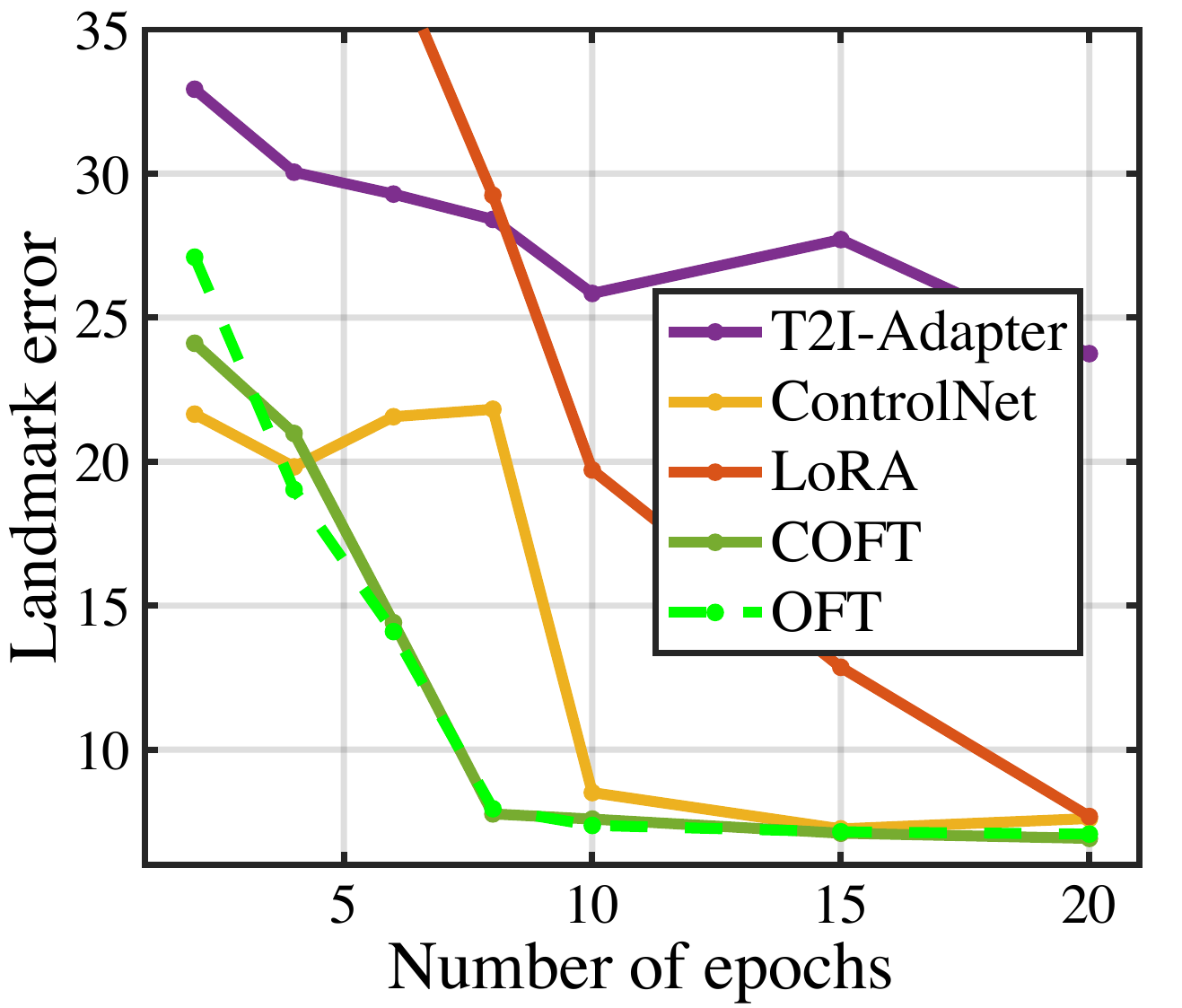}
  \vspace{-1.6em}
    \caption{\scriptsize Face landmark error.}
    \label{fig:face_convergence}
\vspace{-0.5em}
\end{wrapfigure}

\begin{wrapfigure}{r}{0.46\linewidth}
\vspace{-2.3em}
\centering
\includegraphics[width=0.997\linewidth]{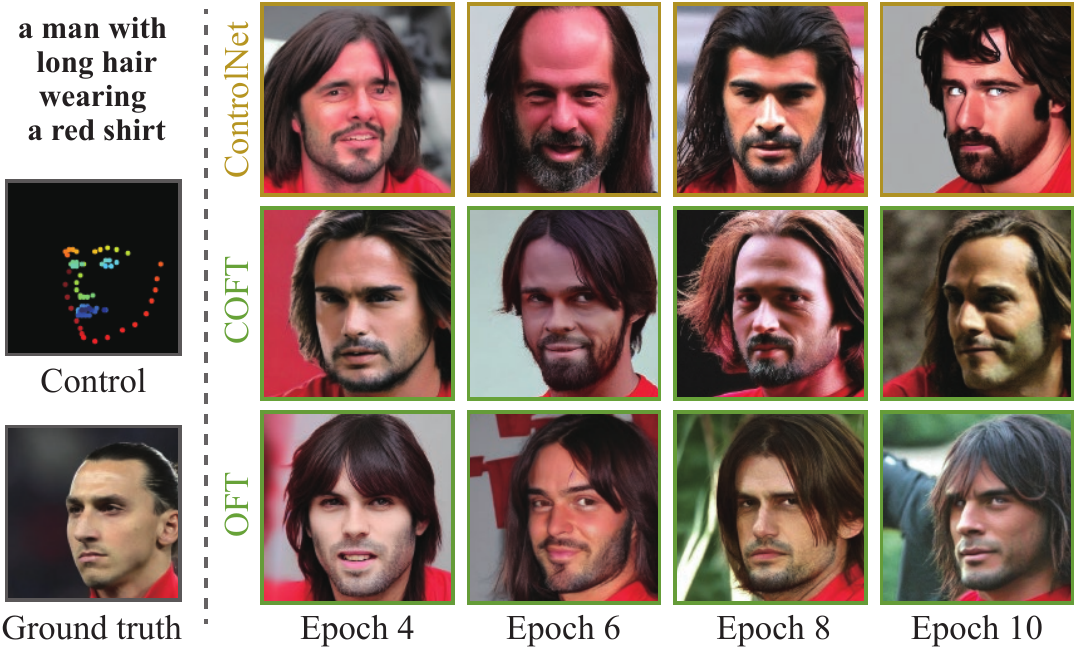}
  \vspace{-1.7em}
    \caption{\scriptsize Qualitative examples with different number of epochs.}
    \label{fig:face_convergence_qual}
\vspace{-0.7em}
\end{wrapfigure}

\textbf{Convergence}. We evaluate the convergence speed of ControlNet, T2I-Adapter, LoRA and COFT on the L2F task. We provide both quantitative and qualitative evaluation. Specifically for the evaluation metric, we compute the mean $\ell_2$ distance between control face landmarks and predicted face landmarks. In Figure~\ref{fig:face_convergence}, we plot the face landmark error obtained by the model finetuned with different number of epochs. We can observe that both COFT and OFT achieve significantly faster convergence. It takes 20 epochs for LoRA to converge to the performance of our OFT framework at the 8-th epoch. We note that OFT and COFT use a similar number of trainable parameters to LoRA (much fewer than ControlNet), while being much more efficient to converge than existing methods. On the other hand, the fast convergence of OFT is also validated by the results in Figure~\ref{fig:teaser}. The right example in Figure~\ref{fig:teaser} shows that OFT is much more data-efficient than ControlNet and LoRA, since OFT can converge well with only 5\% of the ADE20K dataset. For qualitative results, we focus on comparing OFT, COFT and ControlNet, because ControlNet achieves the closest landmark error to ours. Results in Figure~\ref{fig:face_convergence_qual} show that both OFT and COFT converge stably and the generated face pose is gradually aligned with the control landmarks. In contrast to our stable and smooth convergence, the controllability in ControlNet suddenly emerges after the 8-th epoch, which perfectly matches the sudden convergence phenomenon observed in \cite{zhang2023adding}. Such a convergence stability makes our OFT framework much easier to use in practice, since the training dynamics of OFT is far more smooth and predictable. Thus it will be easier to find good OFT's hyperparameters.

\textbf{Quantitative comparison}. We introduce a control consistency metric to evaluate the performance of controllable generation. The basic idea is to compute the control signal from the generated image and then compare it with the original input control signal. For the C2I task, we compute IoU and F1 score. For the S2I task, we compute mean IoU, mean and overall accuracy. For the L2F task, we compute the mean $\ell_2$ distance between control landmarks and predicted landmarks. More details regarding the consistency metrics are given in Appendix~\ref{app:settings}. For all the compared method, we use the best possible hyperparameter settings. Results in Table~\ref{table:control_gen} show that both OFT and COFT yield much stronger and accurate control than the other methods. We observe that the adapter-based approaches (\eg, T2I-Adapter and ControlNet) converge slowly and also yield worse final results. Compared to ControlNet, LoRA performs better in the S2I task and worse in the C2I and L2F tasks. In general, we find that the performance ceiling of LoRA is relatively low, even if we have carefully tuned its hyperparameters. As a comparison, the performance of our OFT framework has not yet saturated, since we empirically find that it still gets better as the number of trainable parameters gets large. We emphasize that our quantitative evaluation in controllable generation is one of our novel contributions, since it can accurately evaluate the control performance of the finetuned models (up to the accuracy of the off-the-shelf segmentation/detection model).

\begin{figure}[t]
    \centering
    \setlength{\abovecaptionskip}{4pt}
    \setlength{\belowcaptionskip}{-8pt}
    \renewcommand{\captionlabelfont}{\scriptsize}
    \includegraphics[width=\textwidth]{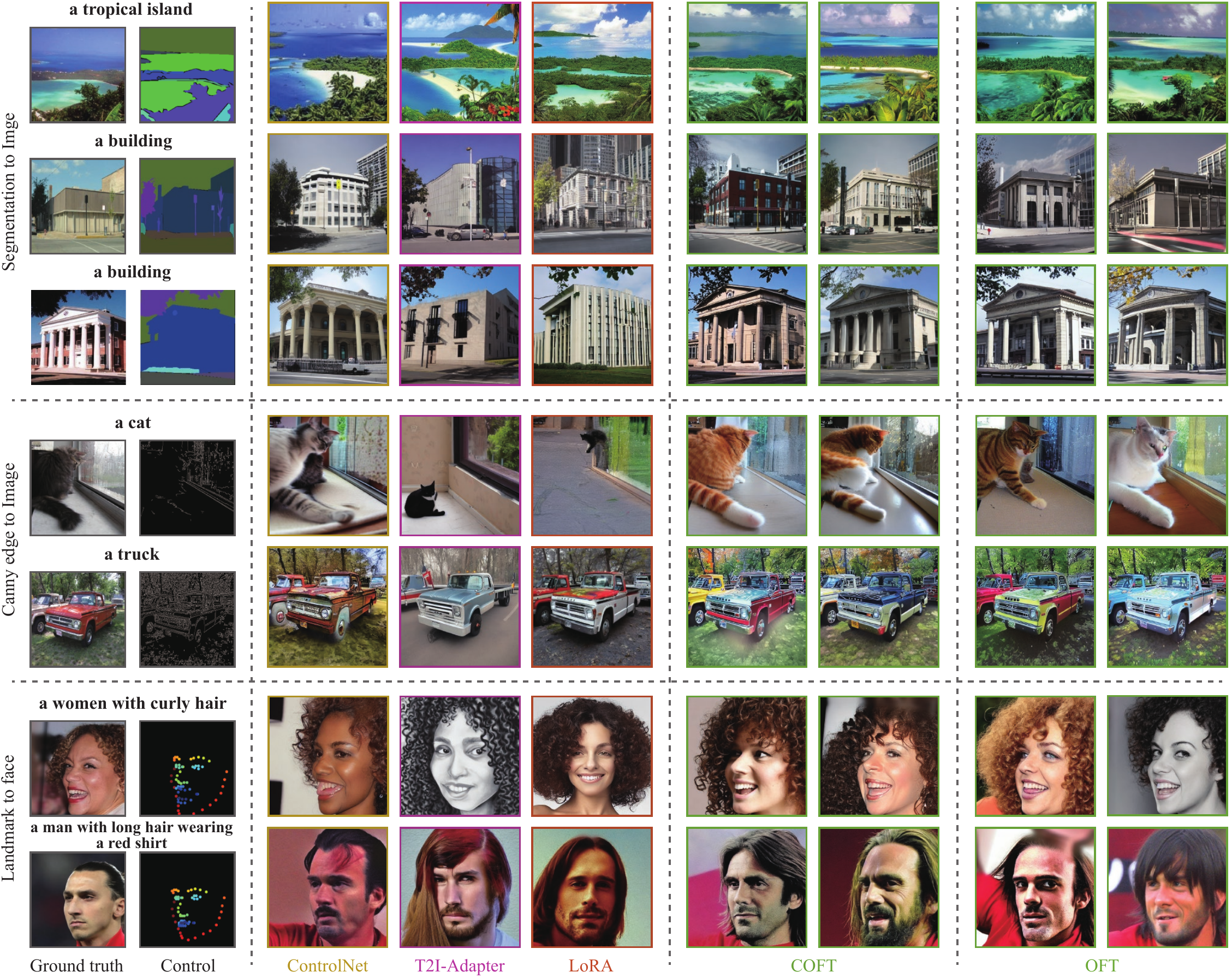}
    \caption{\scriptsize Qualitative comparison of controllable generation. The figure is best viewed digitally, in color and significantly zoomed in.\looseness=-1}
    \label{fig:control_final}
\end{figure}

\setlength{\columnsep}{11pt}
\begin{wraptable}{r}[0cm]{0pt}
    \scriptsize
    \centering
        \hspace{-2.4mm}
    \setlength{\tabcolsep}{3pt}
    \renewcommand{\arraystretch}{1.21}
    \renewcommand{\captionlabelfont}{\footnotesize}
        \begin{tabular}{llcccccc}
        \specialrule{0em}{0pt}{-13.3pt}
        Task & Metric & SD & ControlNet & T2I-Adapter & LoRA &~~COFT~~\cellcolor{Gray}&~~OFT~~\cellcolor{Gray}\\
        \shline
        \multirow{2}{*}[0pt]{C2I} & IoU~$\uparrow$ & 0.049 & 0.189 & 0.078 & 0.168  & \textbf{0.195}\cellcolor{Gray} & 0.193\cellcolor{Gray} \\ 
        &F1~$\uparrow$ & 0.093 & 0.317 & 0.143 & 0.286 & \textbf{0.325}\cellcolor{Gray} & 0.323\cellcolor{Gray} \\\hline 
        \multirow{3}{*}[0pt]{S2I}&mIoU~$\uparrow$ & 7.72 & 20.88 & 16.38 & 22.98 & 26.92\cellcolor{Gray} & \textbf{27.06}\cellcolor{Gray}\\ 
        &mAcc~$\uparrow$ & 14.40 & 30.91 & 26.31 & 35.52 & 40.08\cellcolor{Gray} & \textbf{40.09}\cellcolor{Gray}\\ 
        &aAcc~$\uparrow$ & 33.61 & 61.42 & 51.63 & 58.03 & \textbf{62.96}\cellcolor{Gray} & 62.42\cellcolor{Gray}\\\hline
        L2F&Error~$\downarrow$ & 146.19 & 7.61 & 23.75 & 7.68 & \textbf{6.92}\cellcolor{Gray} & 7.07\cellcolor{Gray}\\
        \specialrule{0em}{-6pt}{0pt}
        \end{tabular}
    \caption{\scriptsize Quantitative comparison of control signal consistency for three control tasks (Canny edge to image, segmentation to image and landmark to face).} \label{table:control_gen}
\vspace{-2.9mm}
\end{wraptable}

\textbf{Qualitative comparison}. We also qualitatively compare OFT and COFT to current state-of-the-art methods, including ControlNet, T2I-Adapter and LoRA. Randomly generated images in Figure~\ref{fig:control_final} show that OFT and COFT not only yield high-fidelity and realistic image quality, but also achieve accurate control. In the S2I task, we can see that LoRA completely fails to generate images following the input segmentation map, while ControlNet, OFT and COFT can well control the generated images. In contrast to ControlNet, both OFT and COFT are able to generate high-fidelity images with more vivid details and more reasonable geometric structures with far less model parameters. In the C2I task, both OFT and COFT are able to hallucinate semantically similar images based on a rough Canny edges, while T2I-Adapter and LoRA perform much worse. In the L2F task, our method produces the most accurate pose control for the generated faces even under challenging face poses. In all three control tasks, we show that both OFT and COFT produce qualitatively better images than the state-of-the-art baselines, demonstrating the effectiveness of our OFT framework in controllable generation. To give a more comprehensive qualitative comparison, we provide more qualitative examples for all the three control tasks in Appendix~\ref{app:control_exp}, and moreover, we demonstrate OFT can perform well on more control tasks (including dense pose to human body, sketch to image and depth to image) in Appendix~\ref{app:more_control_task}.

\vspace{-2mm}
\section{Concluding Remarks and Open Problems}
\vspace{-1.5mm}

Motivated by the observation that angular information among neurons crucially determines visual semantics, we propose a simple yet effective finetuning method – orthogonal finetuning for controlling text-to-image diffusion models. Specifically, we target two text-to-image applications: subject-driven generation and controllable generation. Compared to existing methods, OFT demonstrates stronger controllability and finetuning stability with fewer number of finetuning parameters. More importantly, OFT does not introduce additional inference overhead, leading to an efficient deployable model.

OFT also introduces a few interesting open problems. First, OFT guarantees the orthogonality via Cayley parametrization which involves a matrix inverse. It slightly limits the scalability of OFT. Although we address this limitation using block diagonal parametrization, how to speed up this matrix inverse in a differentiable way remains a challenge. Second, OFT has unique potential in compositionality, in the sense that the orthogonal matrices produced by multiple OFT finetuning tasks can be multiplied together and remains an orthogonal matrix. Whether this set of orthogonal matrices preserve the knowledge of all the downstream tasks remains an interesting direction to study. Finally, the parameter efficiency of OFT is largely dependent on the block diagonal structure which inevitably introduces additional biases and limits the flexibility. How to improve the parameter efficiency in a more effective and less biased way remains an important open problem.

\newpage
\section*{Acknowledgement}
\vspace{-1.5mm}

The authors would like to sincerely thank Luigi Gresele, Yandong Wen, Yuliang Xiu and many other colleagues at Max Planck Institute for Intelligent Systems for many helpful suggestions.

This work was supported by the German Federal Ministry of Education and Research (BMBF): Tubingen AI Center, FKZ: 01IS18039B, and by the Machine Learning Cluster of Excellence, EXC number 2064/1 – Project number 390727645. WL was supported by the German Research Foundation (DFG): SFB 1233, Robust Vision: Inference Principles and Neural Mechanisms, TP XX, project number: 276693517. AW acknowledges support from a Turing AI Fellowship under grant EP/V025279/1, The Alan Turing Institute, and the Leverhulme Trust via CFI.

{
\footnotesize
\bibliographystyle{plain}
\bibliography{ref}
}

\input{new_appendix}

\end{document}

%% file: new_appendix.tex
\clearpage
\newpage

\appendix
\onecolumn

\addcontentsline{toc}{section}{Appendix} 
\renewcommand \thepart{} 
\renewcommand \partname{}
\part{\Large{\centerline{Appendix}}}
\parttoc

\newpage
\section{Experimental Details}\label{app:settings}

To verify the effectiveness of our Orthogonal Fine-tuning (OFT) method, we extensively evaluate the performance of our method in two common text-to-image generation tasks: subject-driven generation and controllable generation. More specifically, we use the exact same task setting as ControlNet~\cite{zhang2023adding} and Dreambooth~\cite{ruiz2023dreambooth} and the baseline implementations were sourced from the GitHub repository Diffusers\footnote{\href{https://github.com/huggingface/diffusers}{https://github.com/huggingface/diffusers}} and ControlNet\footnote{\href{https://github.com/lllyasviel/ControlNet}{https://github.com/lllyasviel/ControlNet}}.

\paragraph{Data and Model.} For training the convolutional autoencoder from Figure~\ref{fig:toy_ae}, we use 1000 random images from the Oxford 102 Flower dataset~\cite{Nilsback08}. For the task of subject-driven generation, we use the official DreamBooth dataset, which consists of 30 subjects from 15 different classes. For each subject, there are several images and 25 different text prompts. For generating the image-control-caption combinations, we use BLIP~\cite{li2022blip} to automatically caption the images (pre-trained model weight and code for captioning based on the GitHub repository BLIP\footnote{\href{https://github.com/salesforce/BLIP}{https://github.com/salesforce/BLIP}}). Note, although COCO provides captions for the training and validation split, to be consistent with other image-control-caption combinations, we instead use the BLIP-generated captions as text prompts. For the C2I task, we use the whole COCO 2017 dataset~\cite{lin2014microsoft} with in total of 180K images; we generate canny edge images as the control signal using the same canny edge detector as ControlNet. For the S2I task, we use the semantic segmentation dataset ADE20K~\cite{zhou2017scene} with in total of 24K image-segmentation mask pairs. For the L2F dataset, we use the CelebA-HQ dataset~\cite{karras2018progressive}, which contains 30K images. Additionally, we demonstrate that OFT also works well in other controllable generation tasks, including depth-to-image (D2I), densepose-to-image (P2I), and sketch-to-image (Sk2I). For the D2I task, we also use the COCO dataset and employ MiDaS~\cite{ranftl2020robust} to generate depth maps; the pre-trained weights are obtained from the GitHub repository MiDaS\footnote{\href{https://github.com/isl-org/MiDaS}{https://github.com/isl-org/MiDaS}}. For the P2I task, we use the DeepFashion-MultiModal dataset~\cite{jiang2022text2human} with in total of 44K clothed human images with the corresponding densepose. For the Sk2I task, we use a subset of the LAION-Aesthetics dataset with approximately 350K images to learn sketch-guided image generation. We use the Stable Diffusion v1.5\footnote{\href{https://huggingface.co/runwayml/stable-diffusion-v1-5/blob/main/v1-5-pruned.ckpt}{https://huggingface.co/runwayml/stable-diffusion-v1-5/blob/main/v1-5-pruned.ckpt}} as the base model.

\paragraph{Subject-driven generation.}

For training our subject-driven generation diffusion model, we follow the training objective of Dreambooth. More specifically, we use the class-specific prior preservation loss to fine-tune our orthogonal matrices:
\begin{equation}
    \label{eq:db_loss}
    \mathbb{E}_{\bm{x},\bm{c},\bm{\epsilon},\bm{\epsilon}^\prime,t}[w_t \|\hat{\bm{x}}_\theta(\alpha_t \bm{x} + \sigma_t \bm{\epsilon}, \bm{c}) - \bm{x} \|^2_2 +
    \lambda w_{t^\prime} \|\hat{\bm{x}}_\theta(\alpha_{t^\prime} \bm{x}_\text{pr} + \sigma_{t^\prime} \bm{\epsilon}^\prime, \bm{c}_\text{pr}) - \bm{x}_\text{pr} \|^2_2],
\end{equation}
with $\bm{c}_\text{pr}$ being the class conditioning vector. For calculating the prior-preservation loss, we additionally need to generate 200 images using the subject's class prompt. Similar to LoRA, we inject our trainable orthogonal matrices into the attention modules of the stable diffusion model. To be comparable with LoRA, we choose the exact same linear layers as LoRA to affect upon: the linear layers $\bm{W}_q$, $\bm{W}_k$, $\bm{W}_v$ and $\bm{W}_o$. We perform training on 1 Tesla V100-SXM2-32GB GPU using a learning rate of $6 \times 10^{-5}$, batch size of 1, and train for approximately 1000 iterations. In the case of COFT, we use $\epsilon = 6 \times 10^{-5}$ to constrain the orthogonal matrices.

\paragraph{Controllable generation.} Apart from injecting the trainable OFT weights into the stable diffusion model, we need to add a small encoding model to stable diffusion to encode the control signal. To be comparable with ControlNet~\cite{zhang2023adding}, we use the same encoding module, which is a shallow 8-layer convolutional network with Scaled Exponential Linear Unit (SELU) activation functions. We also the same training objective as ControlNet. The control signal is encoded and concatenated once with the input to the stable diffusion U-Net. For the LoRA baseline, we use the same encoding module to encode the control signal. For S2I, L2I and P2I, we fine-tune the model for 20 epochs; for C2I and D2I we fine-tune the model for 10 epochs; for Sk2I we fine-tune the model for $8$ epochs. We perform training on $4$ NVIDIA A100-SXM4-80GB GPUs using a learning rate of $1 \times 10^{-5}$, batch size of $4$ for L2I and batch size of $16$ for the rest of tasks. For fine-tuning with COFT, we use $\epsilon = 1 \times 10^{-3}$.

\paragraph{Evaluation.} When evaluating the effectiveness of controllable generation, we primarily focus on evaluating the controllability. Using the consistency metrics introduced in the main paper, we can effectively compute the difference between the control signal and the generated image. For the C2I task, we apply the identical canny filter on the generated image to determine a canny image of the predicted image. Both the control signal canny image and the canny image obtained from the generated images are black-and-white images, with pixel values being either 0 or 1. We evaluate the pixel-wise Intersection over Union (IoU) and F1 score between these two canny predictions. For the S2I task, we compute mean IoU, mean and overall accuracy by deploying a pre-trained semantic segmentation model. More specifically, we use the Segformer\footnote{\href{https://github.com/NVlabs/SegFormer}{https://github.com/NVlabs/SegFormer}}~\cite{xie2021segformer} model, which is trained on ADE20K (Segformer-B4), to perform semantic segmentation on our generated images. We use the segmentation accuracy as an indication for the overall semantically and structural resemblance of the generated images to the ground truth image. 
For the L2F task, we compute the mean $\ell_2$ distance between the input control landmarks and the landmarks estimated from generated images using facial landmark detector~\cite{bulat2017far}.

We also evaluate the generation performance by calculating Fr\'echet Inception Distance (FID)~\cite{heusel2017gans}, we use the default setting of the GitHub repository pytorch-fid\footnote{\href{https://github.com/mseitzer/pytorch-fid}{https://github.com/mseitzer/pytorch-fid}}. The FID is a metric quantifying the similarity between two image dataset. It utilizes 2048-dimensional features, which are derived from  the final average pooling layer of a pretrained InceptionV3 network trained on ImageNet dataset.  A lower FID score indicates a higher similarity between the datasets. 

\newpage
\section{Effect of Different Number of Diagonal Blocks}

We note that the number of diagonal blocks $r$ is an important hyperparameter that effectively controls the number of trainable parameters. It is necessary to perform a sensitivity study on this hyperparameter. Following the same settings as the main paper, we evaluate how $r$ affects OFT in the S2I task. Results in Table~\ref{table:r_study} show that smaller $r$ (closer to recovering the standard orthogonal matrix) generally works better than larger $r$. However, we find that a good trade-off between flexibility and parameter-efficiency indeed exists. Empirically, we find that we can use a much bigger $r$ if the dataset is simple, leading to better parameter-efficiency and faster convergence. In the main paper, we always use $r=4$ because we find that $r=4$ works well across datasets and tasks. Note that, in terms of the number of inference parameters, both LoRA and OFT have the exact same number of parameters, which is equal to the number of parameters of the underlying stable diffusion model, while ControlNet has an additional control model with 361M parameters.

\begin{table}[h]
\centering
\scriptsize
\setlength{\tabcolsep}{6pt}
\renewcommand{\arraystretch}{1.35}
\renewcommand{\captionlabelfont}{\scriptsize}
\begin{tabular}{c c c c c c}
   & \makecell[c]{ControlNet} & \makecell[c]{$r=2$} & \makecell[c]{$r=4$} & \makecell[c]{$r=8$} & \makecell[c]{$r=16$} \\ 
\shline
Trainable Parameters & 361.3 M & 29.5 M & 16.3 M & 9.7 M & 6.4 M\\ 
Inference Parameters & 1.42 B & 1.06 B & 1.06 B & 1.06 B & 1.06 B\\ 
\hline
mIoU~$\uparrow$ & 20.88 & \textbf{27.18} & 27.06 & 24.09 & 21.0\\ 
mAcc~$\uparrow$ & 30.91 & 39.39 & \textbf{40.09} & 36.95 & 32.55\\ 
aAcc~$\uparrow$ & 61.42 & \textbf{65.24} & 62.96 & 60.25 & 55.5\\
\end{tabular}
\vspace{2mm}
\caption{\scriptsize How the number of diagonal blocks affects the control capability of OFT.
}
\vspace{-10pt}
\label{table:r_study}
\end{table}

\newpage
\section{Experiments on Re-scaled OFT}\label{app:rescaled}

Since both OFT and COFT transform neurons with orthogonal matrices and do not affect the magnitude of neurons, their magnitude may be sub-optimal with their updated orientations. To address this issue, we propose a re-scaled OFT where the neuron magnitude is refined using the same set of data in the downstream task. Specifically, re-scaled OFT further finetunes the magnitude of neurons without changing their directions. re-scaled OFT can be performed in two manners: (1) \emph{joint fitting}: magnitude fitting can be used simultaneously with OFT or COFT, and (2) \emph{Post-stage fitting}: magnitude fitting can be used after OFT or COFT is finished. An important motivation for re-scaled OFT comes from Figure~\ref{fig:toy_ae}, where we observe that constructing images only with angular information perfectly preserves visual structures, but it also results in a certain degree of color distortion. We hypothesize that this minor color distortion is caused by magnitude loss and fixing this issue can potentially improve the visual quality of generated images.

Notably, re-scaled OFT does not change the hyperspherical energy since it does not change the direction of neurons - all the nice properties of OFT and COFT on hyperspherical energy are still perfectly preserved. Therefore, the advantage of structural preservation is also inherited.

To simplify the experiments and validate the effectiveness of re-scaled OFT, we perform post-stage magnitude fitting on the COFT model and compare the FID between the original validation images and the generated images (using the control signals extracted from validation images). The reason we use FID here is that FID is more sensitive to color distortion, while the consistency metric only measures the structural preservation. Table~\ref{table:coft+mf} shows that magnitude fitting can indeed improve the FID of COFT and is beneficial to COFT.

Magnitude fitting is lightweight and can be implemented easily by simply adding one trainable parameter for each layer we modify; the parameter has the shape of ($N\times1$), with $N$ corresponds to the number of neurons in that specific layer. The performance gain illustrated in Table~\ref{table:coft+mf} is achieved by performing \emph{Post-stage fitting} on a COFT-fine-tuned model for only one additional epoch. Moreover, we expect that the joint fitting re-scaled OFT can lead to better performance.

\begin{table}[h]
\centering
\scriptsize
\renewcommand{\arraystretch}{1.35}
\setlength{\tabcolsep}{6pt}
\begin{tabular}{c c c c c c c }
   & \makecell[c]{SD} & \makecell[c]{ControlNet} & \makecell[c]{T2I} & \makecell[c]{LoRA} & \makecell[c]{COFT}\cellcolor{Gray} & \makecell[c]{Re-scaled COFT}\cellcolor{Gray} \\ 
\shline   
FID~$\downarrow$ & 41.2 & 30.9 & 33.1 & 30.9 & 30.8\cellcolor{Gray} & \textbf{30.2}\cellcolor{Gray} \\ 
\end{tabular}
\caption{\scriptsize FID on the segmentation to image task (ADE20K). $r=4$ is used here. }
\vspace{-10pt}
\label{table:coft+mf}
\end{table}

\clearpage
\newpage
\section{Applying OFT to Convolution Layers}\label{app:convolution}

In the original setting~\cite{hulora2022}, LoRA is only applied to the linear layers of the attention modules. To be a fair comparison, we also apply OFT to these weights. However, OFT is not limited to linear layers but can easily be adapted to convolution layers by transforming the convolutional neurons. We highlight the compatibility of OFT and COFT for finetuning convolution layers. More interestingly, sharing the parameters of diagonal blocks in $\bm{R}$ becomes interpretable in convolution layers. With a suitable setup, orthogonal matrices with sharing diagonal blocks can transform the convolution kernel in a channel-sharing manner (or in a spatial manner), implying that the same orthogonal transformation is applied to all channels. This shares similar intuition with depth-wise convolution.

\begin{figure}[h]
    \centering
    \setlength{\abovecaptionskip}{6pt}
    \setlength{\belowcaptionskip}{-10pt}
    \renewcommand{\captionlabelfont}{\scriptsize}
    \vspace{0pt}
    \includegraphics[width=0.9\textwidth]{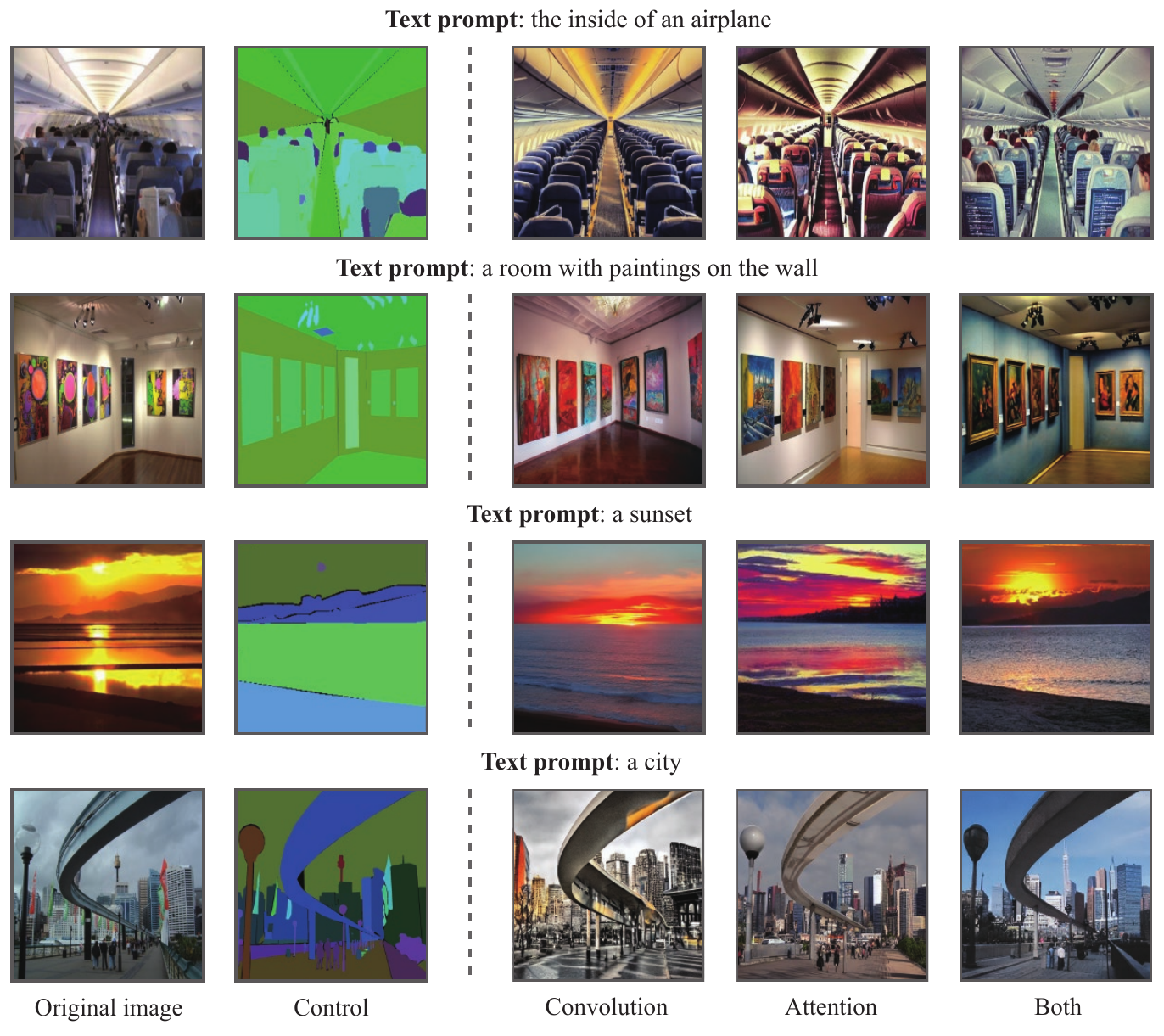}
    \caption{\scriptsize Controllable generation results of applying COFT to different types of layers.\looseness=-1}
    \label{fig:coft_conv}
\end{figure}

For this ablation experiment, we study the performance of applying OFT to the convolution layers in the ResNet blocks of the stable diffusion model. In this experiment, we use COFT as the baseline method and consider the controllable generation (segmentation to image) as an example. We have both quantitative (Table~\ref{table:coft_conv}) and qualitative results (Figure~\ref{fig:coft_conv}). We can empirically observe that by only fine-tuning the convolutional layers, we can also achieve some degree of control. By simultaneously fine-tuning both linear and convolutional layers, we achieve a slightly better FID score. Note, for fine-tuning convolutional layers, we let $r$ be equal to the number of channels of convolutional neurons in that layer.

\begin{table}[h]
\centering
\scriptsize
\renewcommand{\arraystretch}{1.35}
\setlength{\tabcolsep}{6pt}
\begin{tabular}{c c c c}
   & \makecell[c]{COFT (attention)} & \makecell[c]{COFT (conv)} & \makecell[c]{COFT (extended)} \\ 
\shline   
FID~$\downarrow$ & 30.8 & 39.8 & \textbf{30.4} \\ 
\end{tabular}
\caption{\scriptsize FID results of applying COFT to different types of layers. (with $r=4$)}
\vspace{-10pt}
\label{table:coft_conv}
\end{table}

\clearpage
\newpage

\section{Comparison between COFT and OFT}\label{app:coft_oft}
\begin{figure}[h]
    \centering
    \setlength{\abovecaptionskip}{6pt}
    \setlength{\belowcaptionskip}{-18pt}
    \renewcommand{\captionlabelfont}{\scriptsize}
    \vspace{-2.5pt}
    \includegraphics[width=0.97\textwidth]{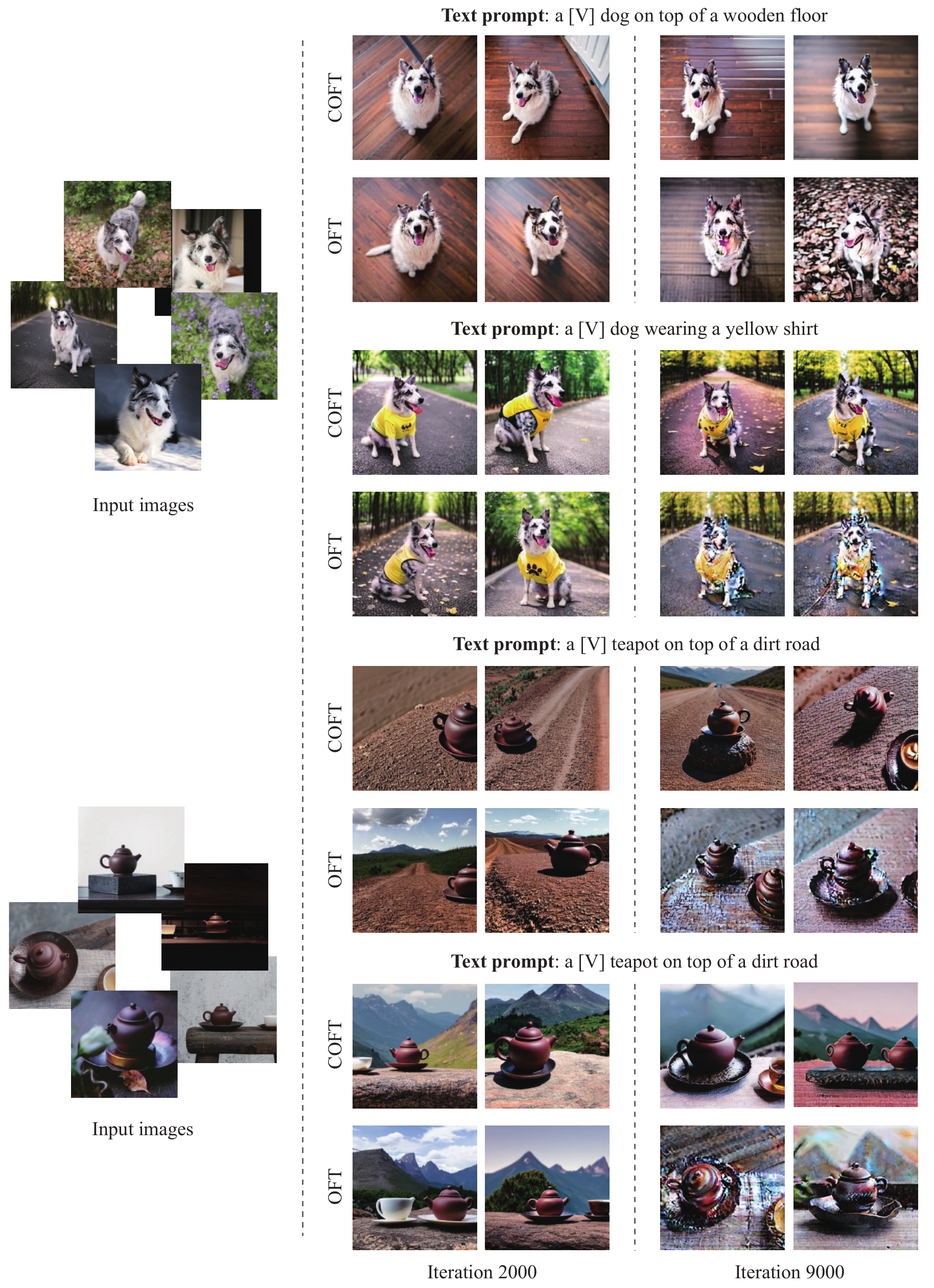}
    \caption{\scriptsize Qualitative comparison between COFT and OFT on subject-driven generation.\looseness=-1}
    \label{fig:oft_vs_coft1}
\end{figure}

\begin{figure}[h]
    \centering
    \setlength{\abovecaptionskip}{8pt}
    \setlength{\belowcaptionskip}{-10pt}
    \renewcommand{\captionlabelfont}{\scriptsize}
    \vspace{0pt}
    \includegraphics[width=0.99\textwidth]{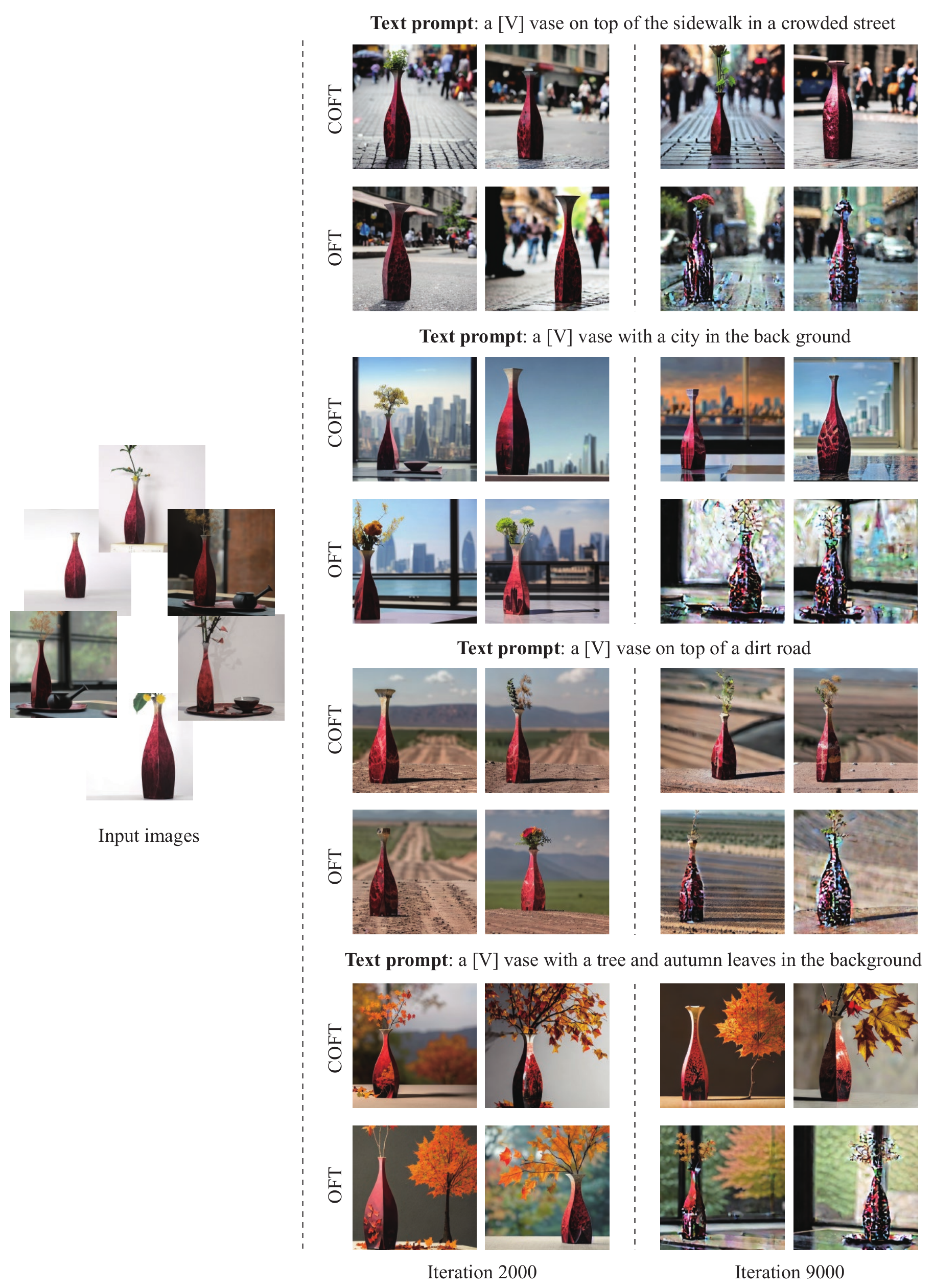}
    \caption{\scriptsize Qualitative comparison between COFT and OFT on subject-driven generation.\looseness=-1}
    \label{fig:oft_vs_coft2}
\end{figure}

\vspace{-1mm}
We have already provided many qualitative examples for COFT and OFT in the main paper. One may question the fundamental difference between OFT and COFT. Based on the intuition behind COFT, the deviation constraint is introduced to improve the training stability. We demonstrate the training stability of COFT with a qualitative example in subject-driven generation. Results in Figure~\ref{fig:oft_vs_coft1} and Figure~\ref{fig:oft_vs_coft2} show that, despite being much more stable than existing methods, OFT will eventually generate collapsed images at the 9000-th iteration. In contrast, COFT still produces visually appealing images. We train both OFT and COFT with a learning rate of $1 \times 10^{-5}$ and constrain COFT with $\epsilon = 1 \times 10^{-5}$.

\clearpage
\newpage
\section{More Qualitative Results}\label{app:more_qualitative}

\subsection{Subject-driven Generation}

\begin{figure}[h]
    \centering
    \setlength{\abovecaptionskip}{4pt}
    \setlength{\belowcaptionskip}{-10pt}
    \renewcommand{\captionlabelfont}{\scriptsize}
    \vspace{0pt}
    \includegraphics[width=0.97\textwidth]{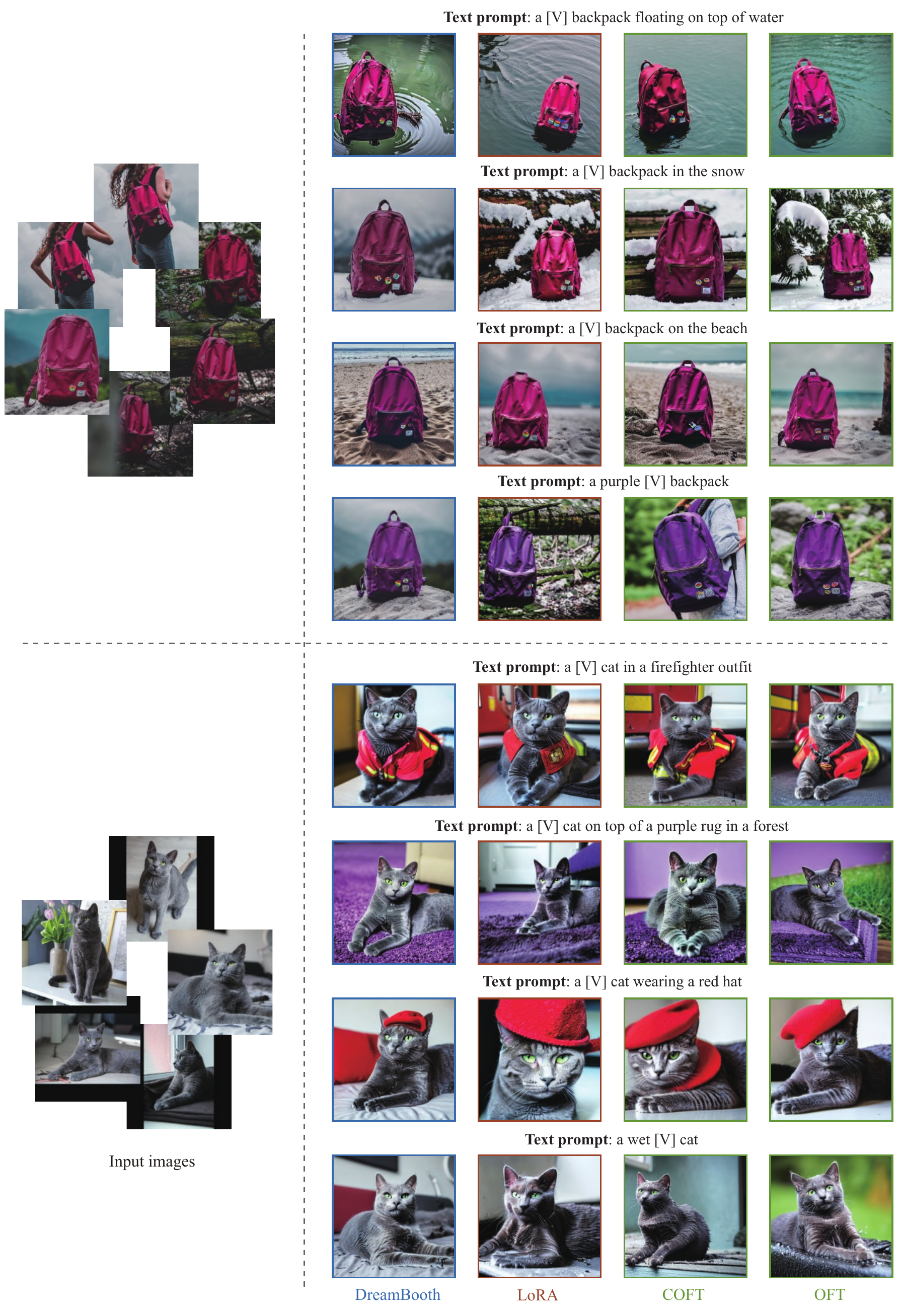}
    \caption{\scriptsize More qualitative results on subject-driven generation.\looseness=-1}
    \label{fig:db_app}
\end{figure}

\begin{figure}[h]
    \centering
    \setlength{\abovecaptionskip}{4pt}
    \setlength{\belowcaptionskip}{-10pt}
    \renewcommand{\captionlabelfont}{\scriptsize}
    \vspace{0pt}
    \includegraphics[width=0.97\textwidth]{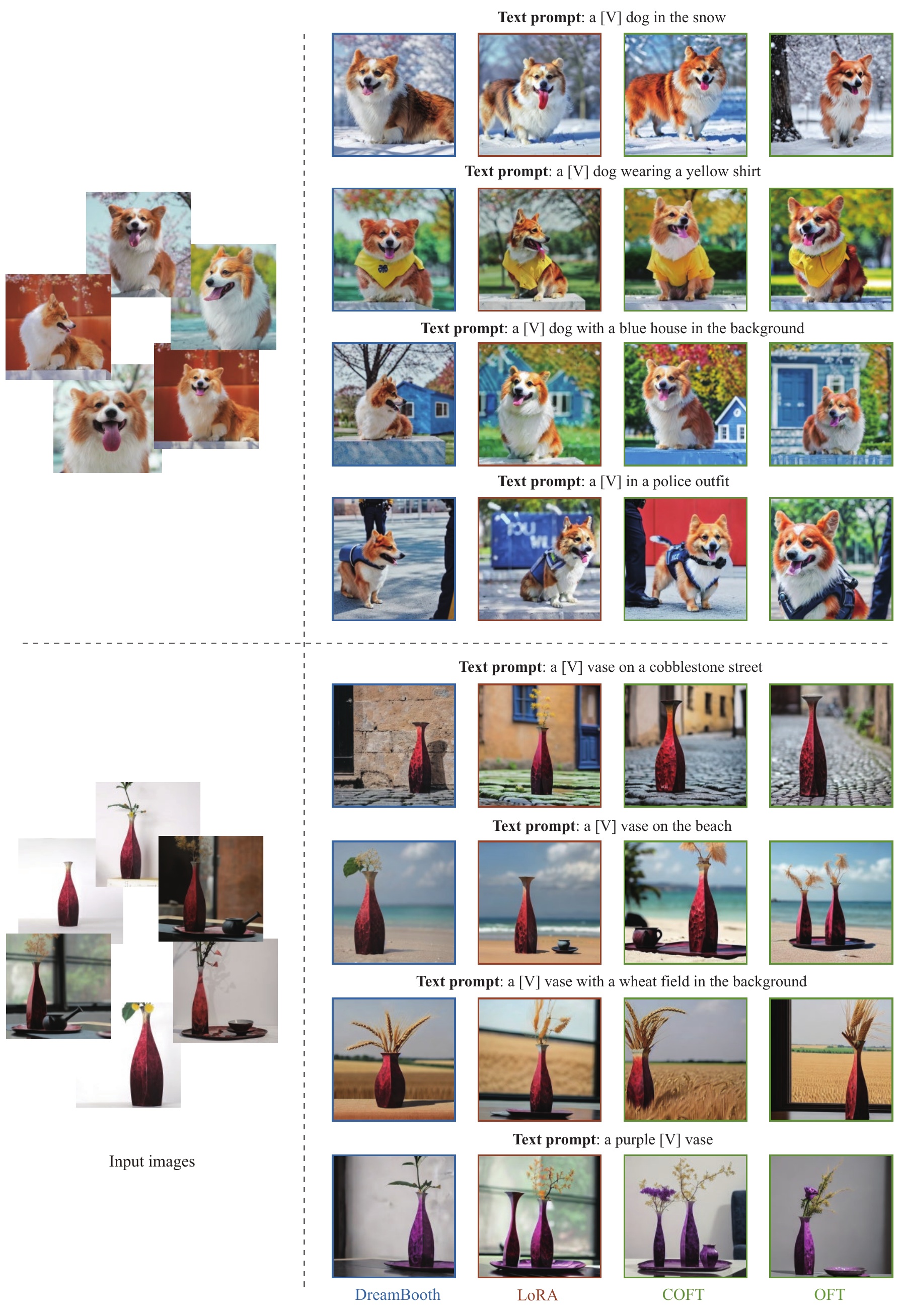}
    \caption{\scriptsize More qualitative results on subject-driven generation.\looseness=-1}
    \label{fig:db_app2}
\end{figure}

\clearpage
\newpage
\subsection{Controllable Generation}\label{app:control_exp}

\subsubsection{Segmentation to Image}
\begin{figure}[h]
    \centering
    \setlength{\abovecaptionskip}{4pt}
    \setlength{\belowcaptionskip}{-27pt}
    \renewcommand{\captionlabelfont}{\scriptsize}
    \vspace{-5pt}
    \includegraphics[width=.97\textwidth]{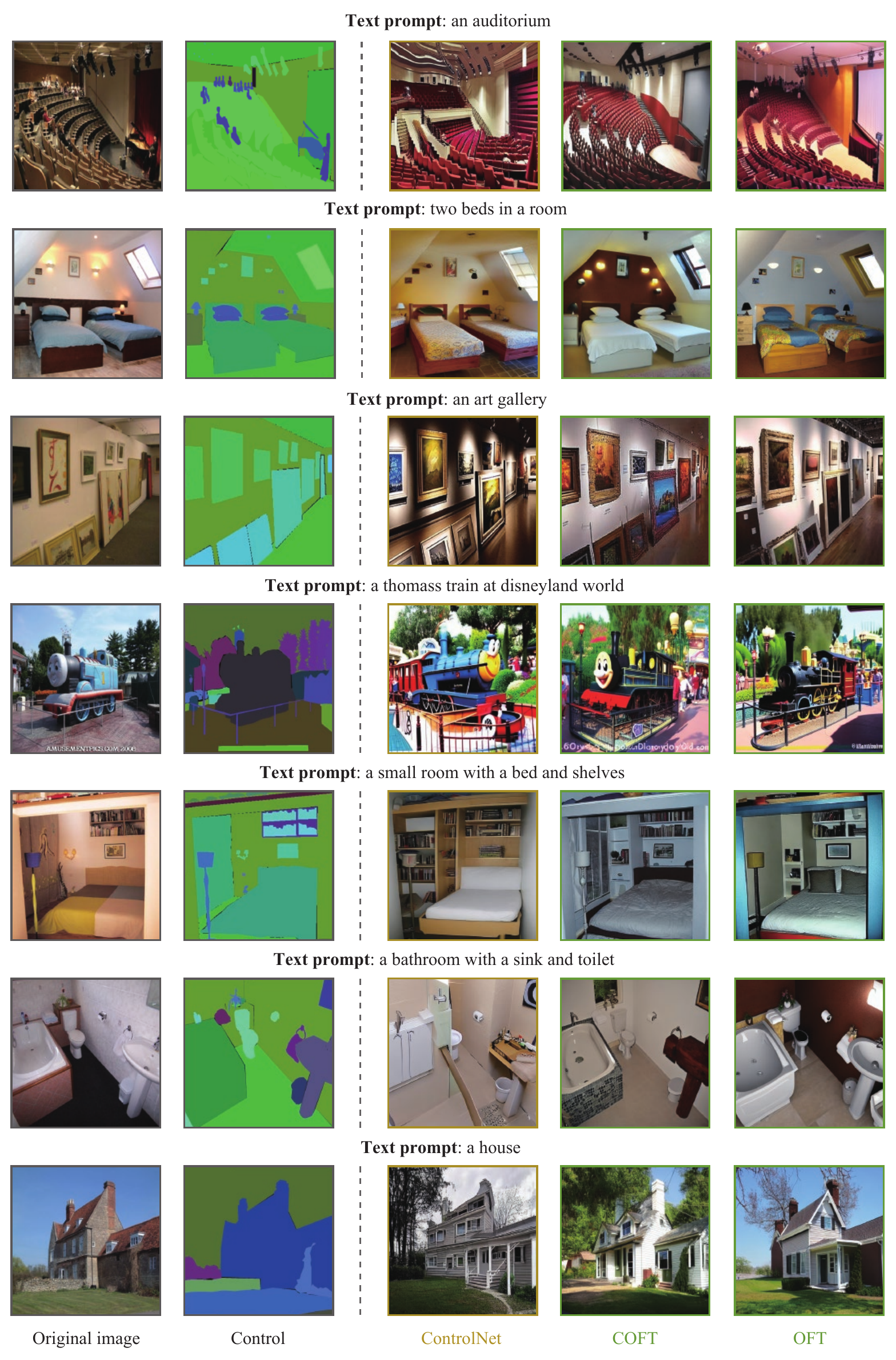}
    \caption{\scriptsize More qualitative results of OFT and COFT on the segmentation to image generation task.\looseness=-1}
    \label{fig:seg2img_app}
\end{figure}

\begin{figure}[h]
    \centering
    \setlength{\abovecaptionskip}{4pt}
    \setlength{\belowcaptionskip}{-27pt}
    \renewcommand{\captionlabelfont}{\scriptsize}
    \vspace{-5pt}
    \includegraphics[width=.97\textwidth]{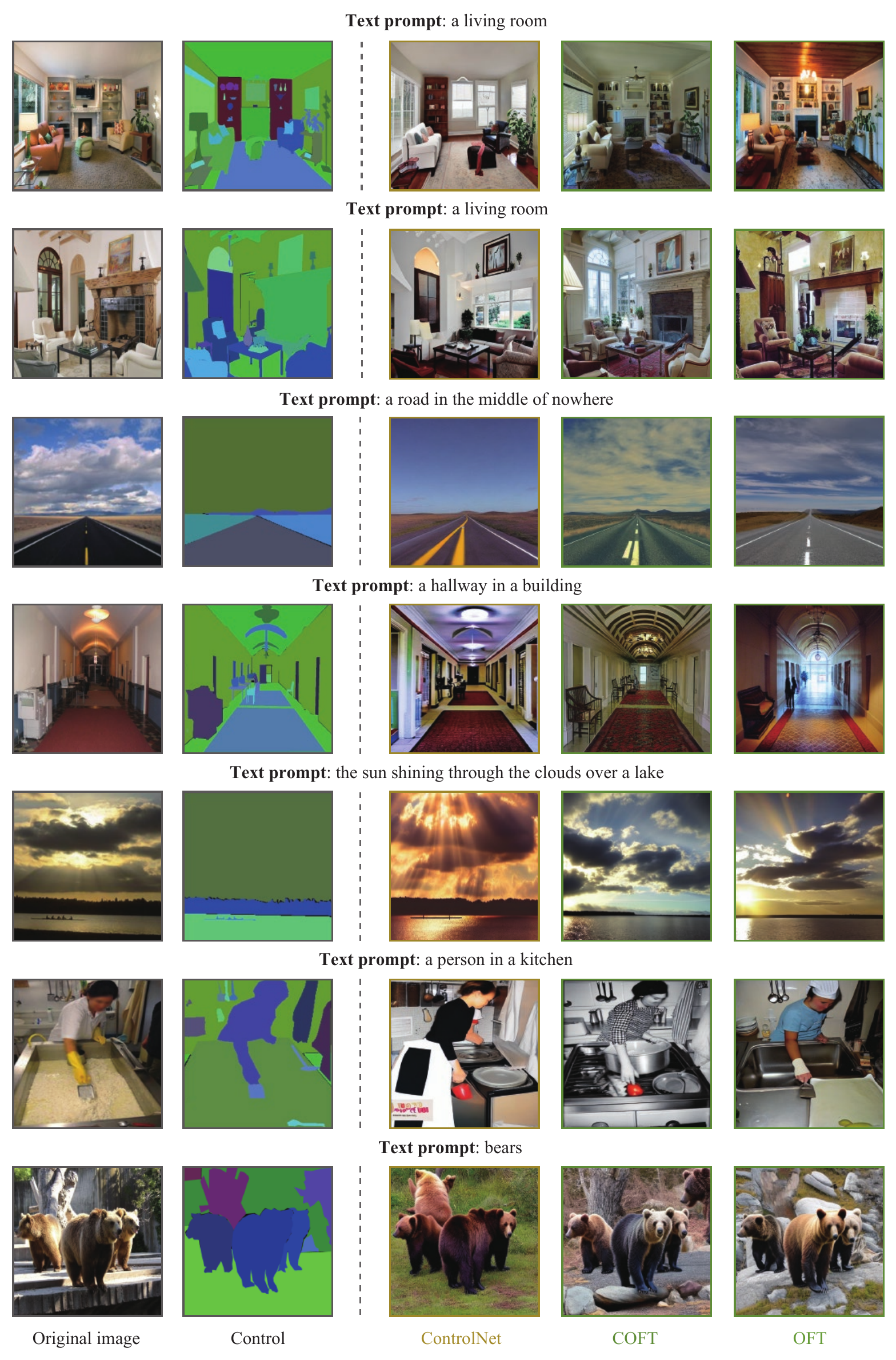}
    \caption{\scriptsize More qualitative results of OFT and COFT on the segmentation to image generation task.\looseness=-1}
    \label{fig:seg2img_app2}
\end{figure}

\begin{figure}[h]
    \centering
    \setlength{\abovecaptionskip}{4pt}
    \setlength{\belowcaptionskip}{-27pt}
    \renewcommand{\captionlabelfont}{\scriptsize}
    \vspace{-5pt}
    \includegraphics[width=.97\textwidth]{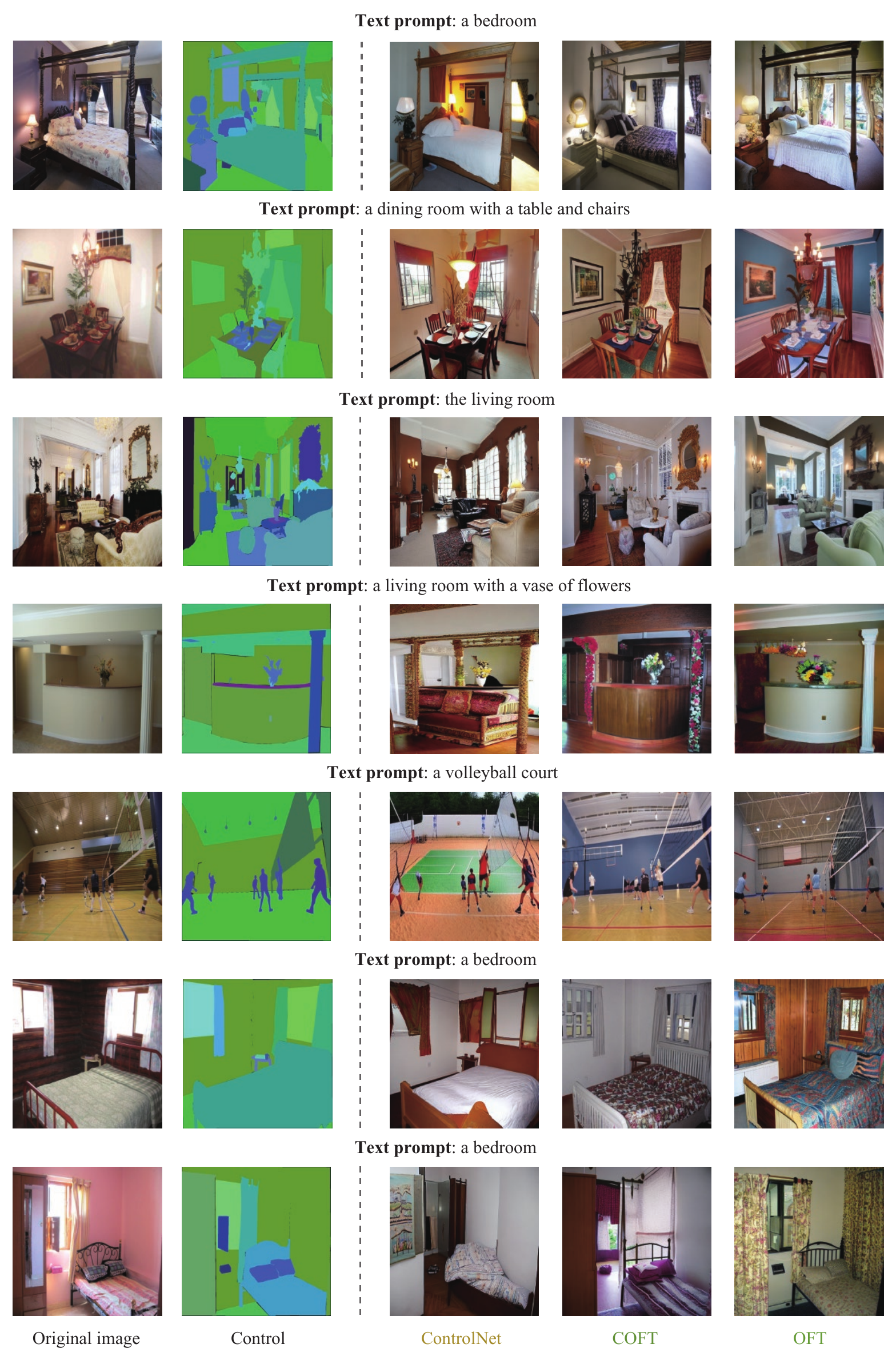}
    \caption{\scriptsize More qualitative results of OFT and COFT on the segmentation to image generation task.\looseness=-1}
    \label{fig:seg2img_app3}
\end{figure}

\clearpage
\newpage
\subsubsection{Canny Edge to Image}

\begin{figure}[h]
    \centering
    \setlength{\abovecaptionskip}{4pt}
    \setlength{\belowcaptionskip}{-27pt}
    \renewcommand{\captionlabelfont}{\scriptsize}
    \vspace{-5pt}
    \includegraphics[width=.97\textwidth]{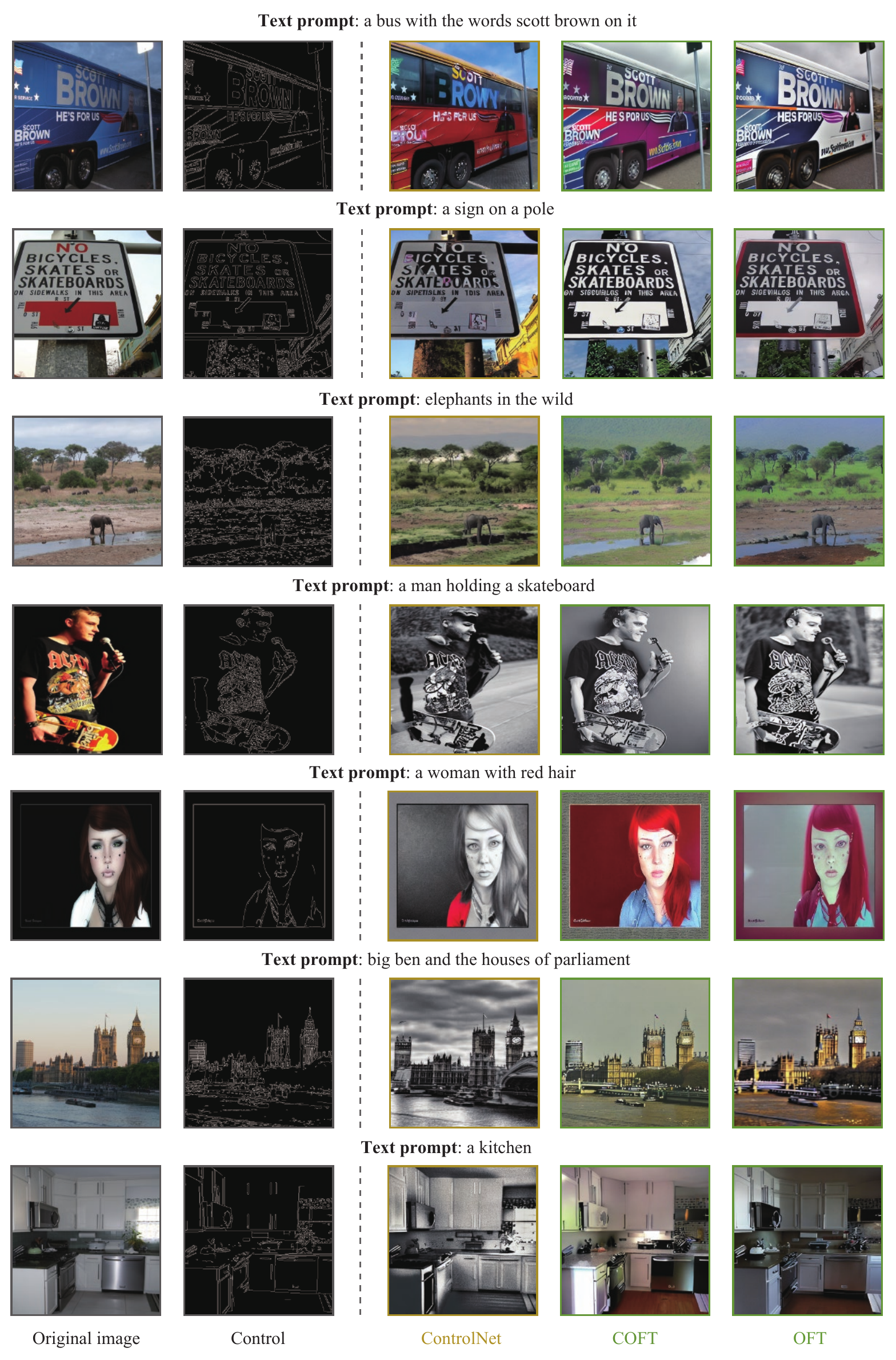}
    \caption{\scriptsize More qualitative results of OFT and COFT on the Canny edge to image generation task.\looseness=-1}
    \label{fig:canny2img_app}
\end{figure}

\begin{figure}[h]
    \centering
    \setlength{\abovecaptionskip}{4pt}
    \setlength{\belowcaptionskip}{-27pt}
    \renewcommand{\captionlabelfont}{\scriptsize}
    \vspace{-5pt}
    \includegraphics[width=.97\textwidth]{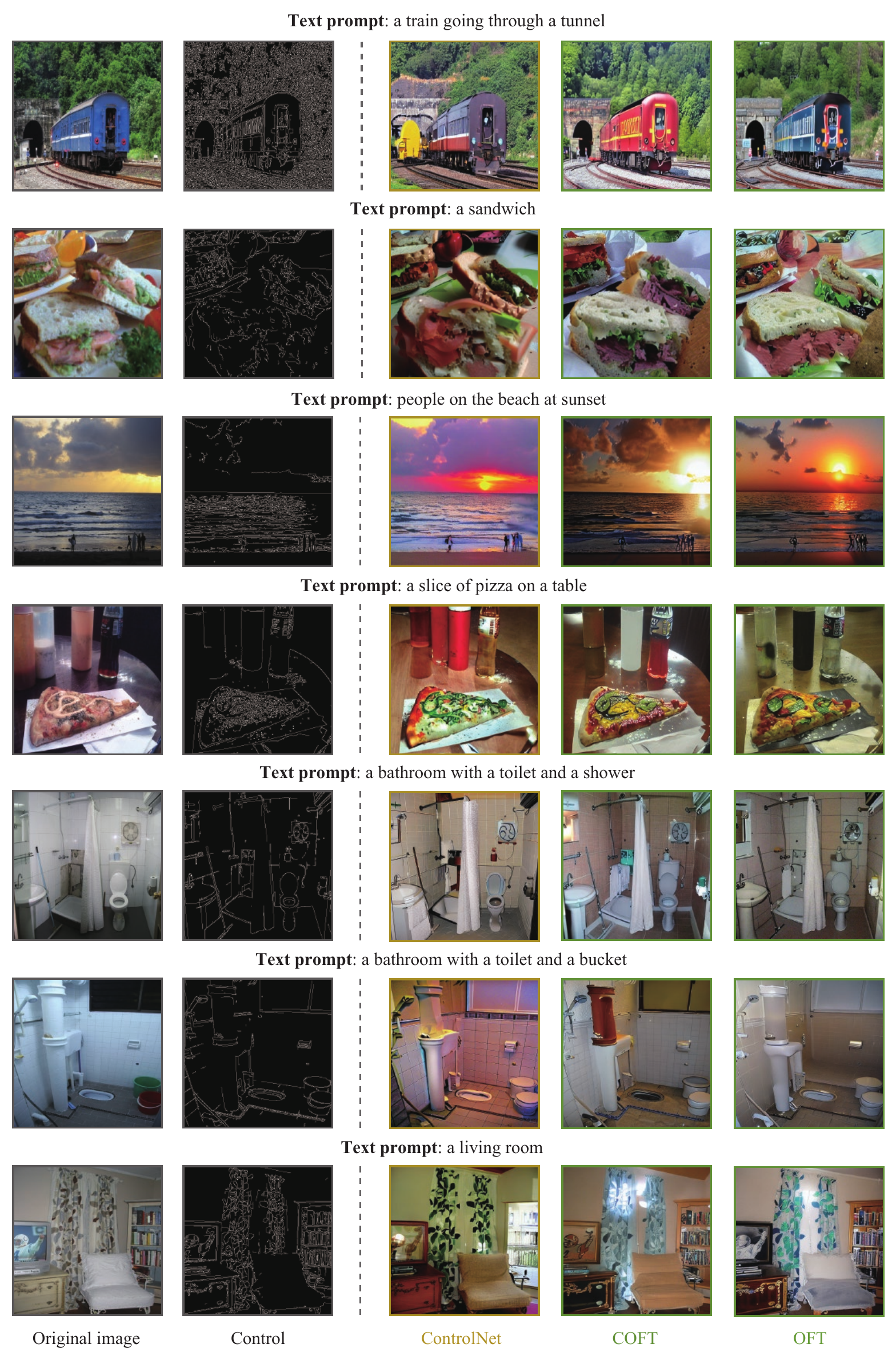}
    \caption{\scriptsize More qualitative results of OFT and COFT on the Canny edge to image generation task.\looseness=-1}
    \label{fig:canny2img_app2}
\end{figure}

\begin{figure}[h]
    \centering
    \setlength{\abovecaptionskip}{4pt}
    \setlength{\belowcaptionskip}{-27pt}
    \renewcommand{\captionlabelfont}{\scriptsize}
    \vspace{-5pt}
    \includegraphics[width=.97\textwidth]{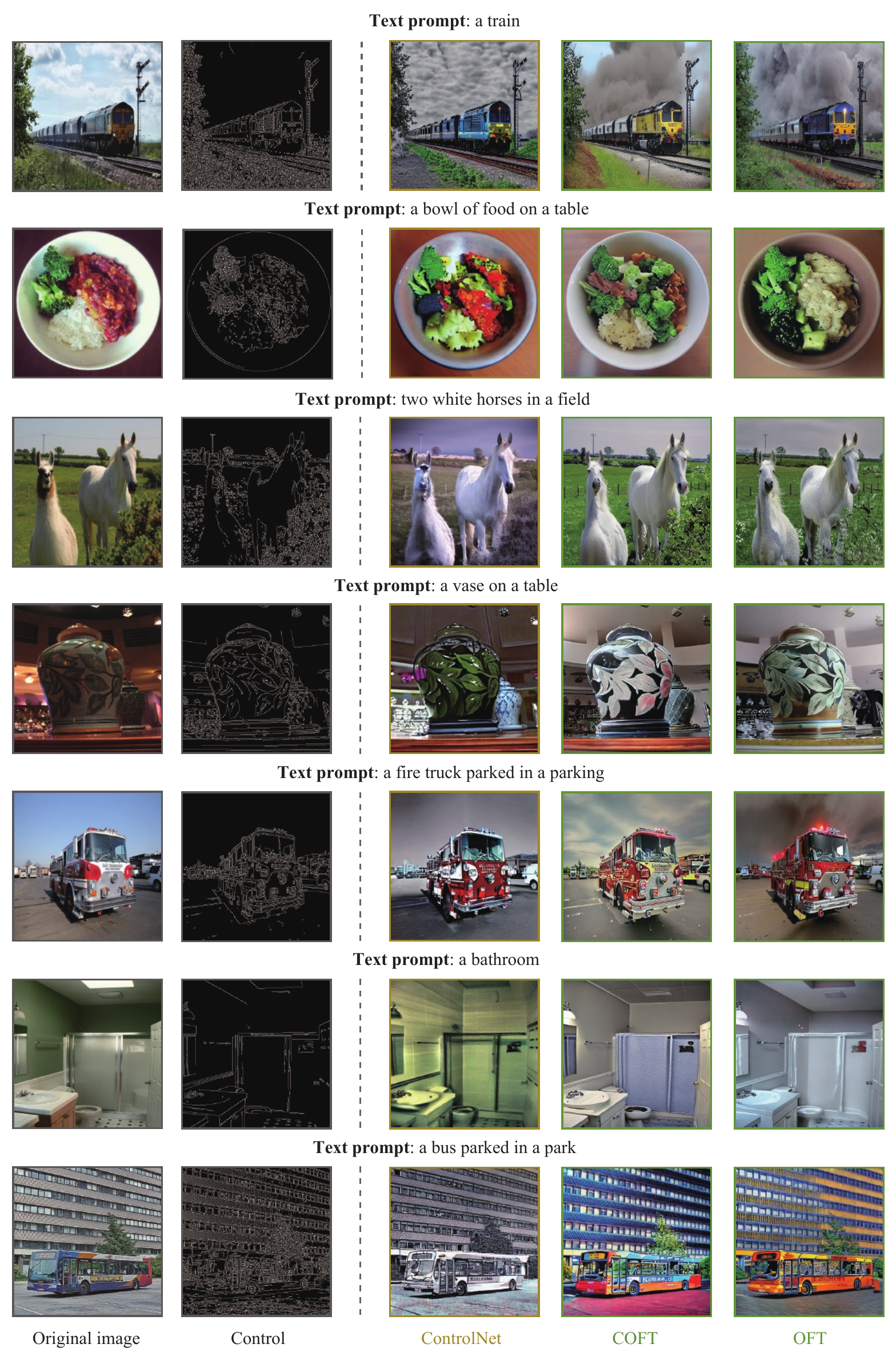}
    \caption{\scriptsize More qualitative results of OFT and COFT on the Canny edge to image generation task.\looseness=-1}
    \label{fig:canny2img_app3}
\end{figure}

\clearpage
\newpage
\subsubsection{Landmark to Face}

\begin{figure}[h]
    \centering
    \setlength{\abovecaptionskip}{4pt}
    \setlength{\belowcaptionskip}{-27pt}
    \renewcommand{\captionlabelfont}{\scriptsize}
    \vspace{-5pt}
    \includegraphics[width=.97\textwidth]{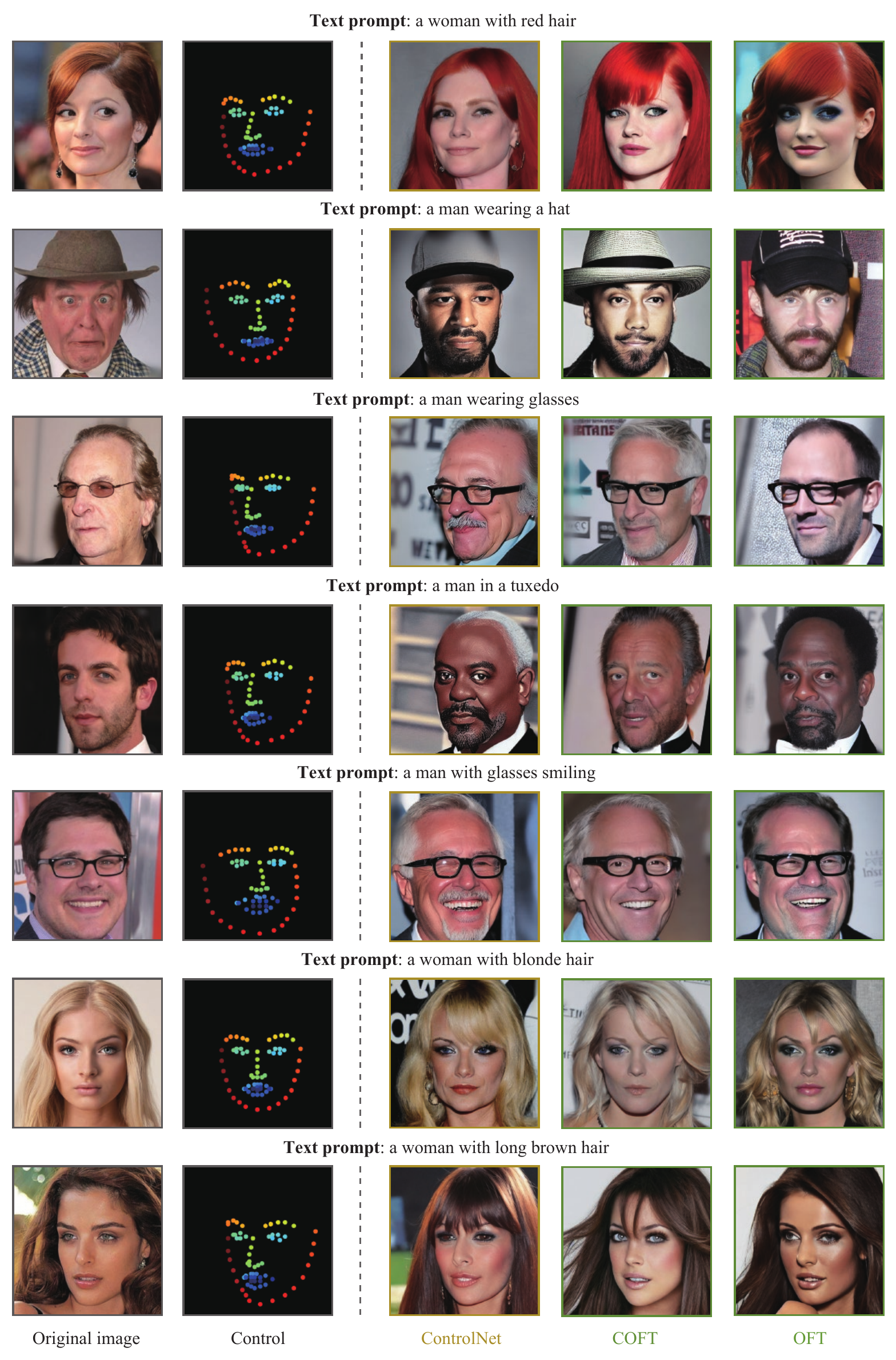}
    \caption{\scriptsize More qualitative results of OFT and COFT on the landmark to face generation task.\looseness=-1}
    \label{fig:land2face_app}
\end{figure}

\begin{figure}[h]
    \centering
    \setlength{\abovecaptionskip}{4pt}
    \setlength{\belowcaptionskip}{-27pt}
    \renewcommand{\captionlabelfont}{\scriptsize}
    \vspace{-5pt}
    \includegraphics[width=.97\textwidth]{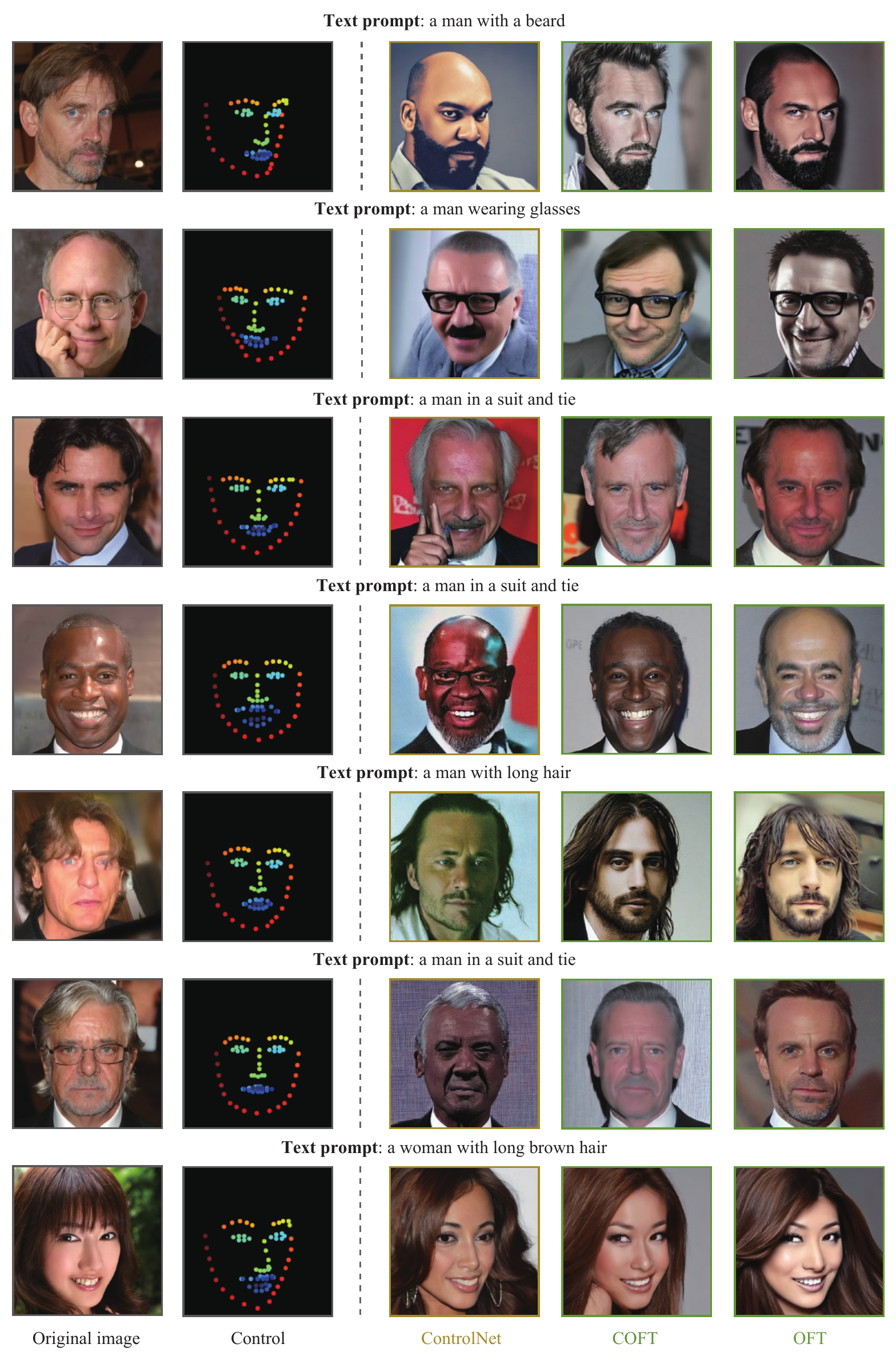}
    \caption{\scriptsize More qualitative results of OFT and COFT on the landmark to face generation task.\looseness=-1}
    \label{fig:land2face_app2}
\end{figure}

\begin{figure}[h]
    \centering
    \setlength{\abovecaptionskip}{4pt}
    \setlength{\belowcaptionskip}{-27pt}
    \renewcommand{\captionlabelfont}{\scriptsize}
    \vspace{-5pt}
    \includegraphics[width=.97\textwidth]{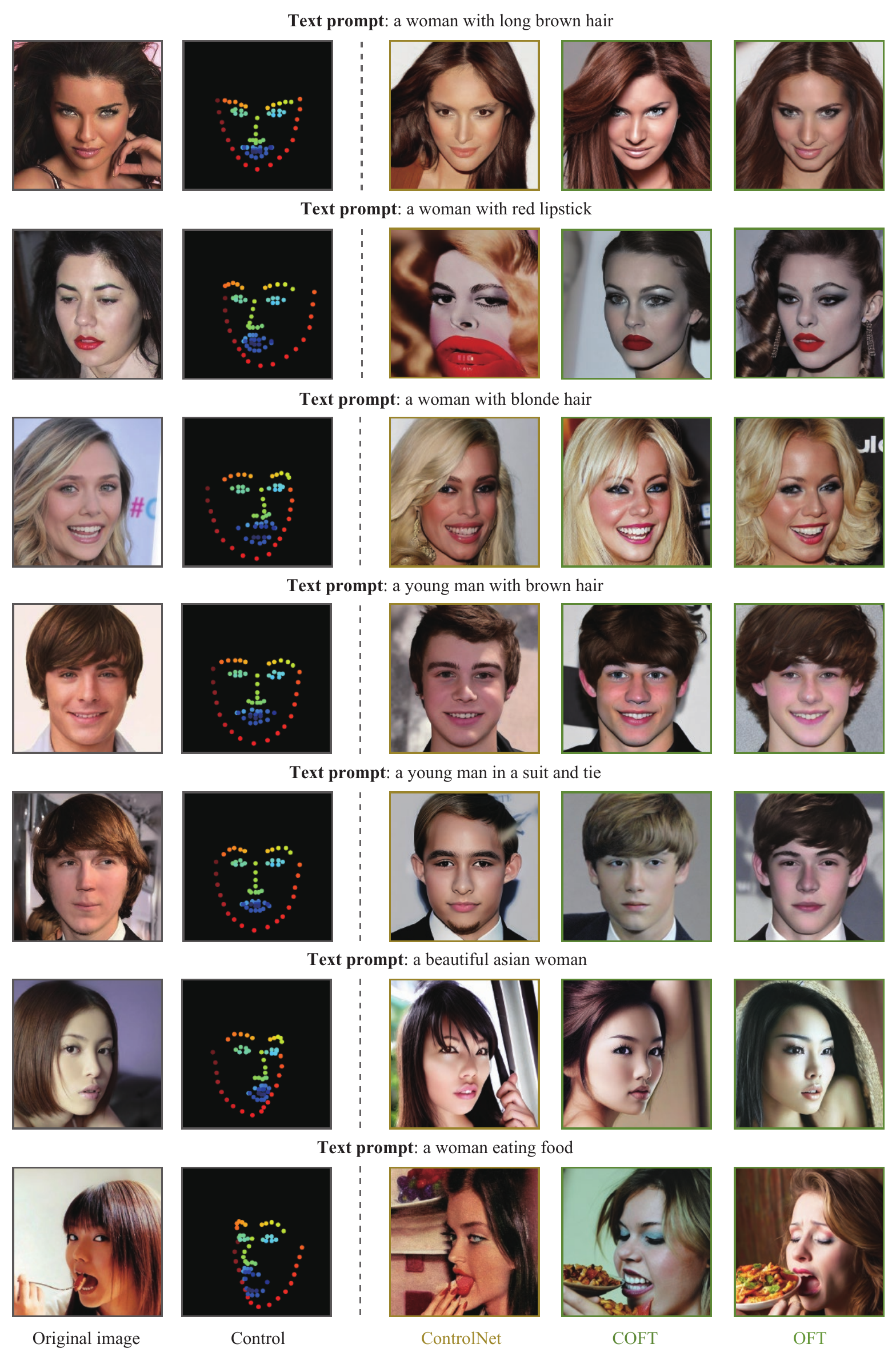}
    \caption{\scriptsize More qualitative results of OFT and COFT on the landmark to face generation task.\looseness=-1}
    \label{fig:land2face_app3}
\end{figure}

\clearpage
\newpage
\section{More Controllable Generation Tasks}\label{app:more_control_task}
\vspace{-0.75mm}
\subsection{Dense Pose to Human Body}

\begin{figure}[h]
    \centering
    \setlength{\abovecaptionskip}{4pt}
    \setlength{\belowcaptionskip}{-27pt}
    \renewcommand{\captionlabelfont}{\scriptsize}
    \vspace{-5pt}
    \includegraphics[width=\textwidth]{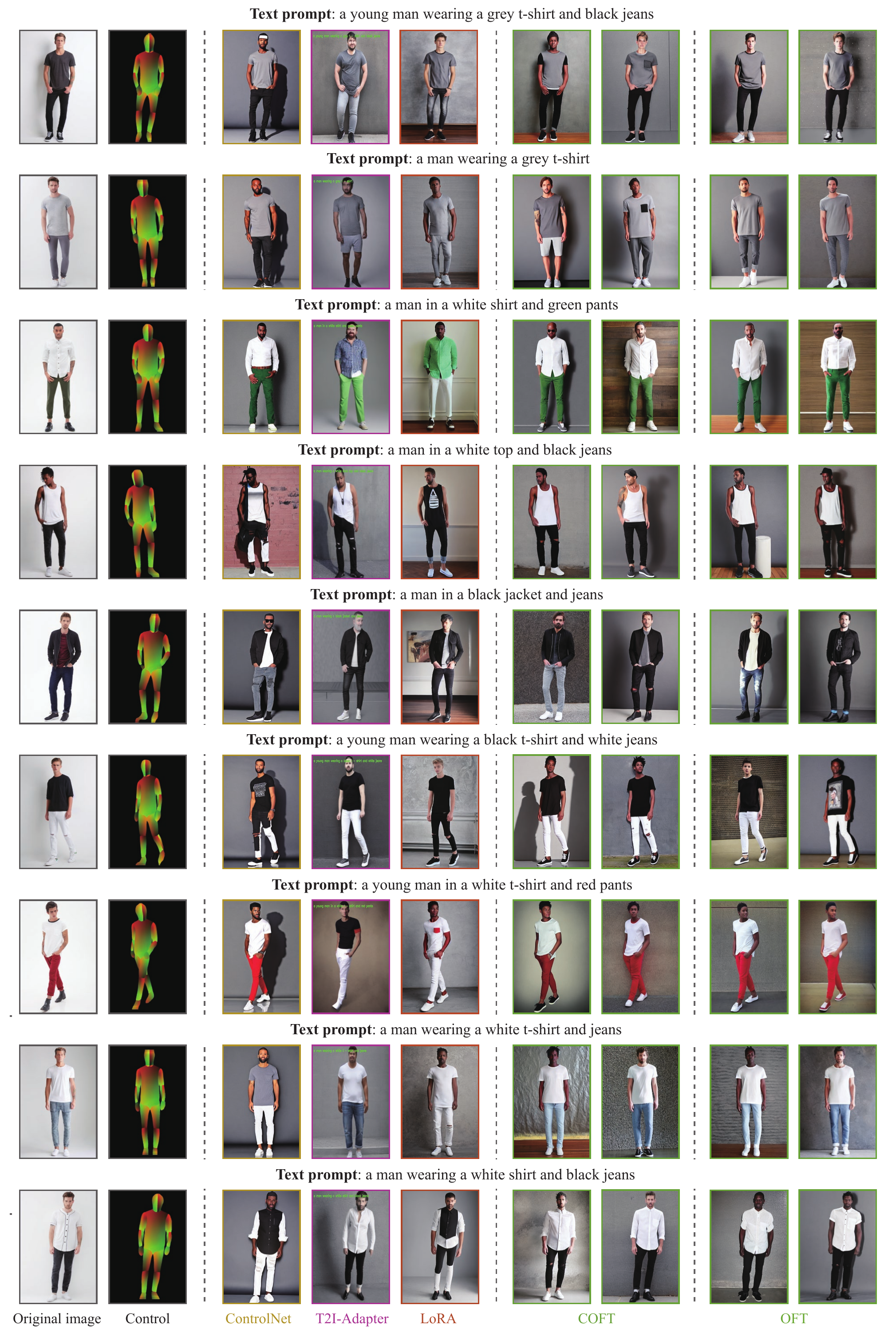}
    \caption{\scriptsize Qualitative comparison among different methods on the dense pose to human body generation task.\looseness=-1}
    \label{fig:pose2body}
\end{figure}

\begin{figure}[h]
    \centering
    \setlength{\abovecaptionskip}{4pt}
    \setlength{\belowcaptionskip}{-10pt}
    \renewcommand{\captionlabelfont}{\scriptsize}
    \vspace{-5pt}
    \includegraphics[width=\textwidth]{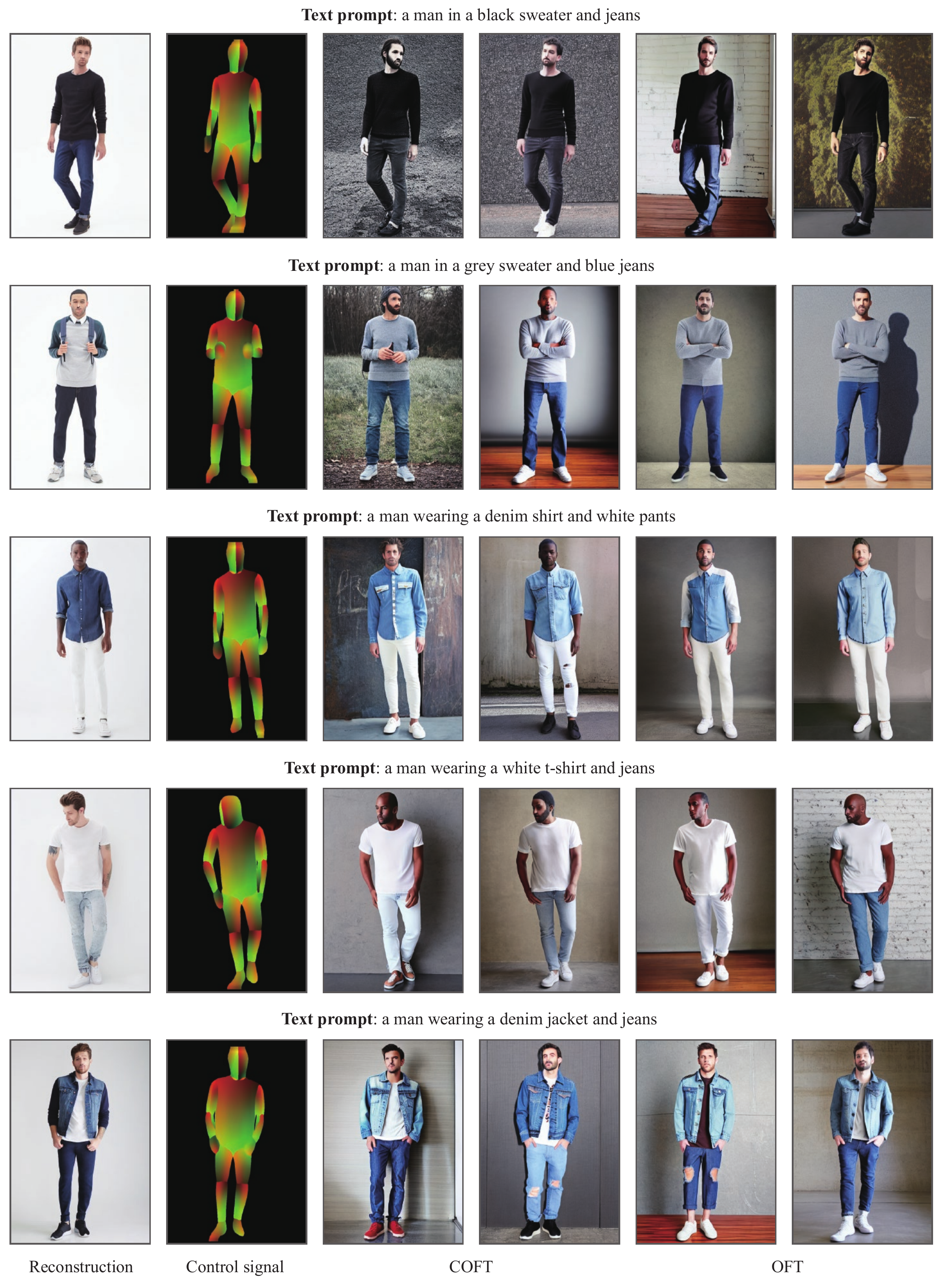}
    \caption{\scriptsize More qualitative results of COFT and OFT on the dense pose to human body task.\looseness=-1}
    \label{fig:pose2body2}
\end{figure}

\clearpage
\newpage
\subsection{Sketch to Image}
\begin{figure}[h]
    \centering
    \setlength{\abovecaptionskip}{4pt}
    \setlength{\belowcaptionskip}{-10pt}
    \renewcommand{\captionlabelfont}{\scriptsize}
    \vspace{-2pt}
    \includegraphics[width=\textwidth]{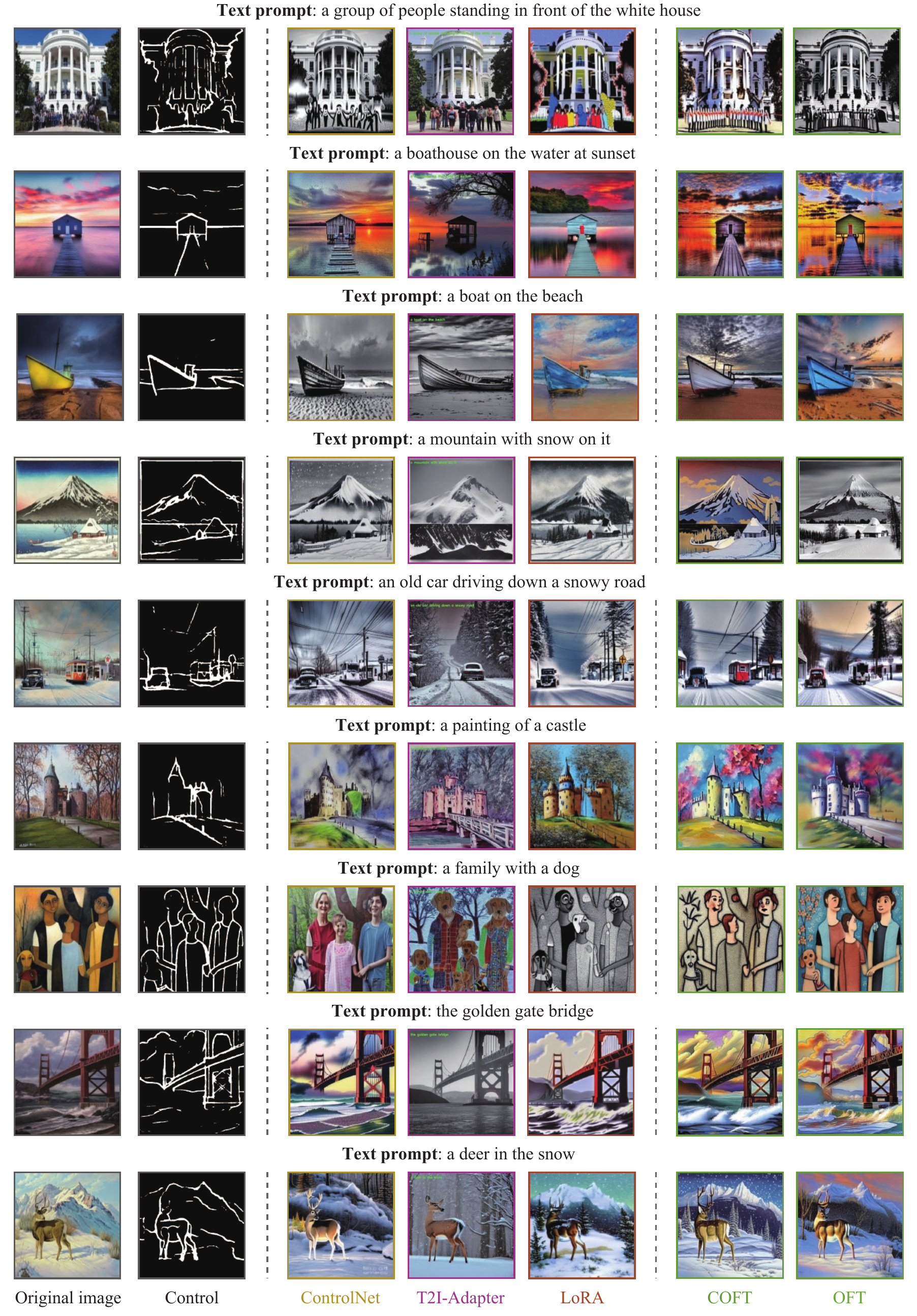}
    \caption{\scriptsize Qualitative comparison among different methods on the sketch to image generation task.\looseness=-1}
    \label{fig:ske2img}
\end{figure}

\newpage
\begin{figure}[h]
    \centering
    \setlength{\abovecaptionskip}{4pt}
    \setlength{\belowcaptionskip}{-10pt}
    \renewcommand{\captionlabelfont}{\scriptsize}
    \vspace{-5pt}
    \includegraphics[width=\textwidth]{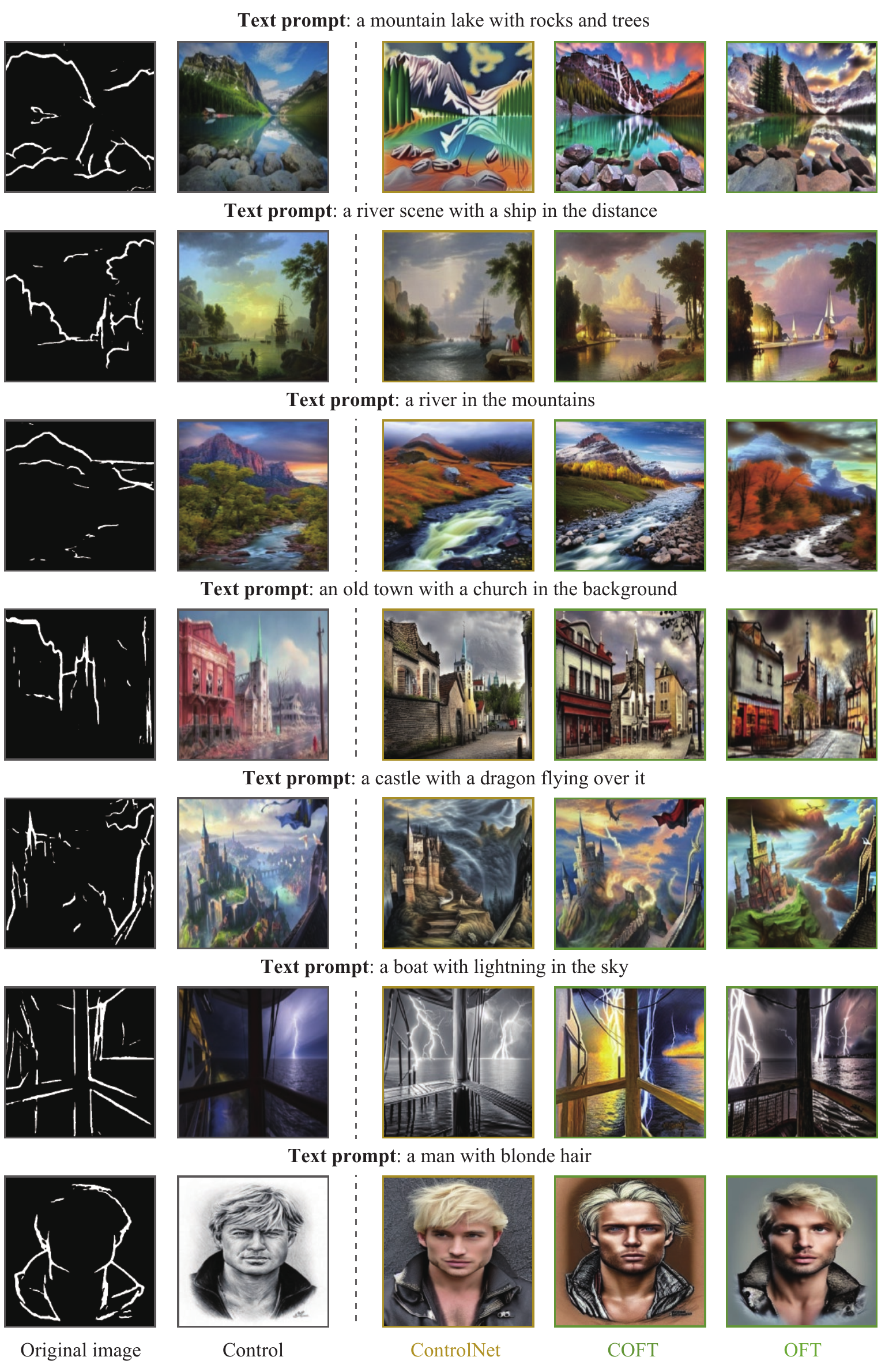}
    \caption{\scriptsize More qualitative comparison on the sketch to image generation task.\looseness=-1}
    \label{fig:ske2img2}
\end{figure}

\clearpage
\newpage
\subsection{Depth to Image}
\begin{figure}[h]
    \centering
    \setlength{\abovecaptionskip}{4pt}
    \setlength{\belowcaptionskip}{-10pt}
    \renewcommand{\captionlabelfont}{\scriptsize}
    \vspace{-2pt}
    \includegraphics[width=\textwidth]{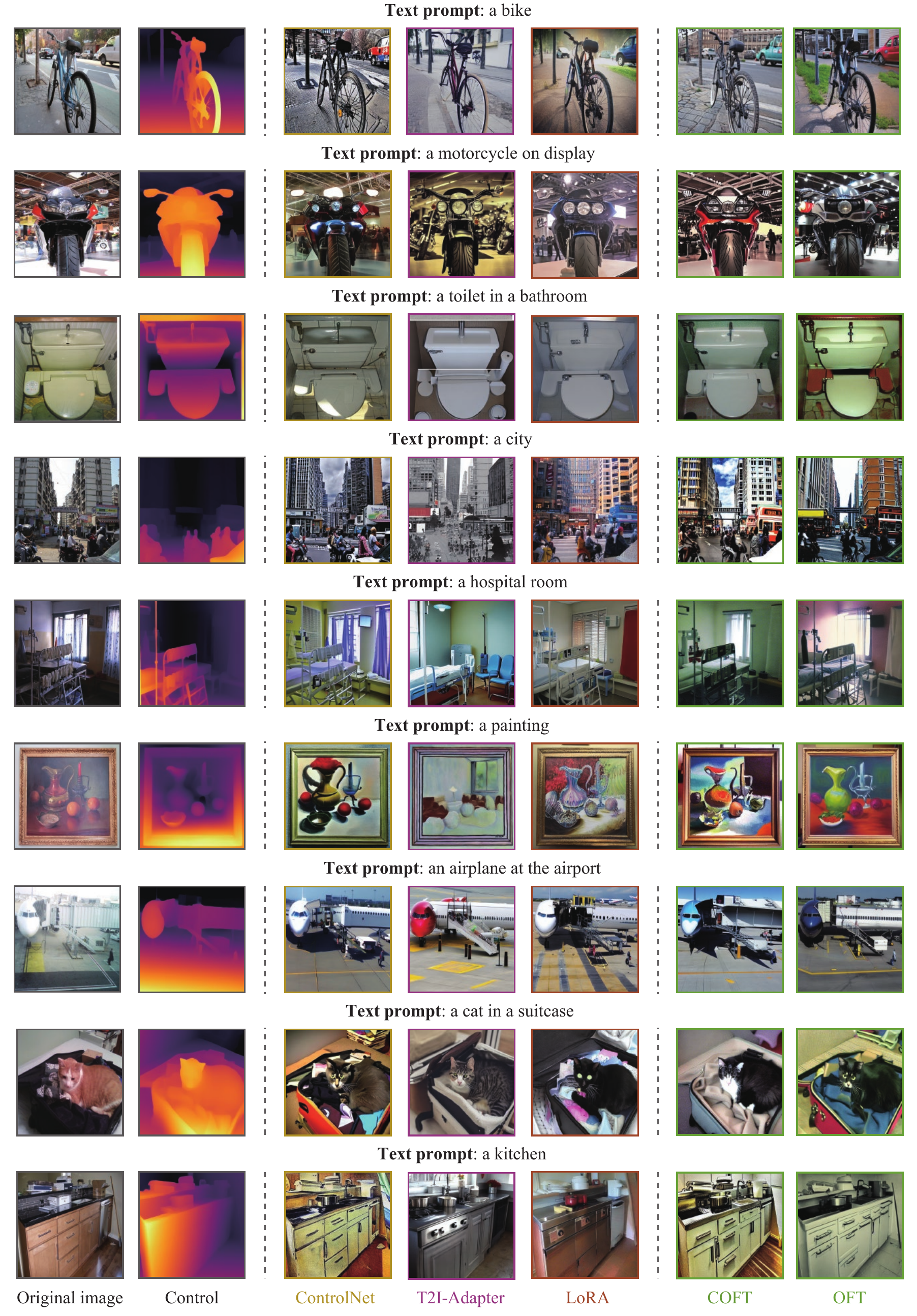}
    \caption{\scriptsize Qualitative comparison among different methods on the depth to image generation task.\looseness=-1}
    \label{fig:depth2img}
\end{figure}

\newpage
\begin{figure}[h]
    \centering
    \setlength{\abovecaptionskip}{4pt}
    \setlength{\belowcaptionskip}{-10pt}
    \renewcommand{\captionlabelfont}{\scriptsize}
    \vspace{-5pt}
    \includegraphics[width=\textwidth]{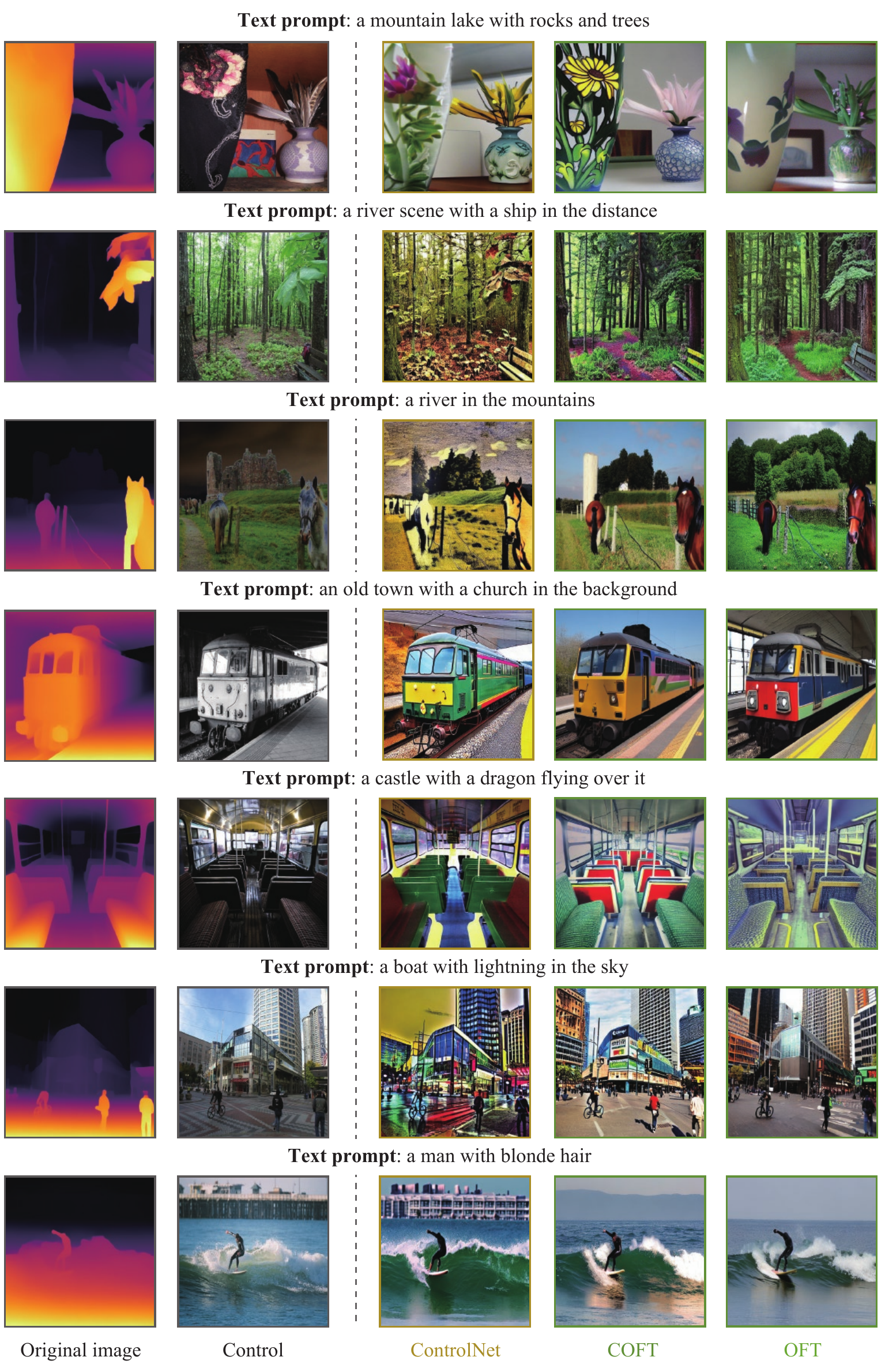}
    \caption{\scriptsize More qualitative comparison on the depth to image generation task.\looseness=-1}
    \label{fig:depth2img2}
\end{figure}

\clearpage
\newpage
\section{Human Evaluation}\label{app:human}


\textbf{Human evaluation settings}. We also carried out a structured human evaluation for the \textbf{subject-driven generation} task, involving 50 participants. Here's a breakdown of our evaluation process:
\begin{itemize}[topsep=0pt, leftmargin=*]
\item \textbf{Selection of subjects}: we picked 7 subjects from the DreamBooth dataset\footnote{\href{https://github.com/google/dreambooth}{https://github.com/google/dreambooth}} at random.
\item \textbf{Image and prompt}: for every subject, 4 unique text prompts were chosen at random. This resulted in a total of 28 distinct subject-prompt combinations. For every single one of the 28 tasks, we randomly sampled an image generated by each of the three methods - DreamBooth, LoRA, and OFT.
\end{itemize}
Every participant was asked to answer three single-selection questions for each task:
\begin{itemize}[topsep=0pt, leftmargin=*]
\item \textbf{Subject fidelity}: which image best preserves the identity of the subject? In other words, which generated image resembles the most the original subject?
\item \textbf{Text alignment}: which image matches the given text description the best?
\item \textbf{Overall image quality}: out of the options, which image has the best overall quality?
\end{itemize}

The methods were assessed at two specific points during their fine-tuning phase: at the \textbf{1000th} iteration, a checkpoint where these methods typically exhibit best performance, and at the \textbf{10,000th} iteration, a checkpoint used to measure the stability of the finetuning process over an extended period.

\textbf{Results}. The results are given in Table~\ref{table:human_eval}, indicating the proportion of participants who chose a particular method based on the above criteria. We can see that OFT is more favored after finetuning Stable Diffusion with 1000 iterations and after 10000 iterations. We note that OFT delivers significantly better image quality and text-following ability than both DreamBooth and LoRA after a relatively large number of finetuning iterations.

\begin{table}[h]
	\centering
\scriptsize
\renewcommand{\arraystretch}{1.35}
\setlength{\tabcolsep}{6pt}
\begin{tabular}{c|ccc|ccc} 
  & \multicolumn{3}{c|}{Iteration 1000} & \multicolumn{3}{c}{Iteration 10000} \\
Metric & DreamBooth & LoRA & \cellcolor{Gray}OFT & DreamBooth & LoRA & \cellcolor{Gray}OFT\\ \shline
Subject Fidelity & 42.0\% & 15.4\% &\cellcolor{Gray}\bf  42.6\% & 22.4\% & 1.4\% & \cellcolor{Gray}\bf 76.2\%\\
Text Alignment & 18.6\% & 24.7\% &\cellcolor{Gray}\bf  56.7\% & 2.6\% & 1.4\% &\cellcolor{Gray}\bf  96.0\%\\
Overall Image Quality & 35.7\% & 19.2\% &\cellcolor{Gray}\bf  45.1\% & 11.6\% & 0.8\% &\cellcolor{Gray}\bf  87.6\%\\
\specialrule{0em}{0pt}{1pt}
\end{tabular}
\caption{\scriptsize Participant voting percentages for subject fidelity, text alignment and overall image quality.}
\label{table:human_eval}
\vspace{-2mm}
\end{table}

\newpage
\section{Style Transfer by Adapting Stable Diffusion with Orthogonal Finetuning}

\begin{figure}[h]
    \centering
    \setlength{\abovecaptionskip}{10pt}
    \setlength{\belowcaptionskip}{0pt}
    \vspace{-2.5pt}
    \includegraphics[width=\textwidth]{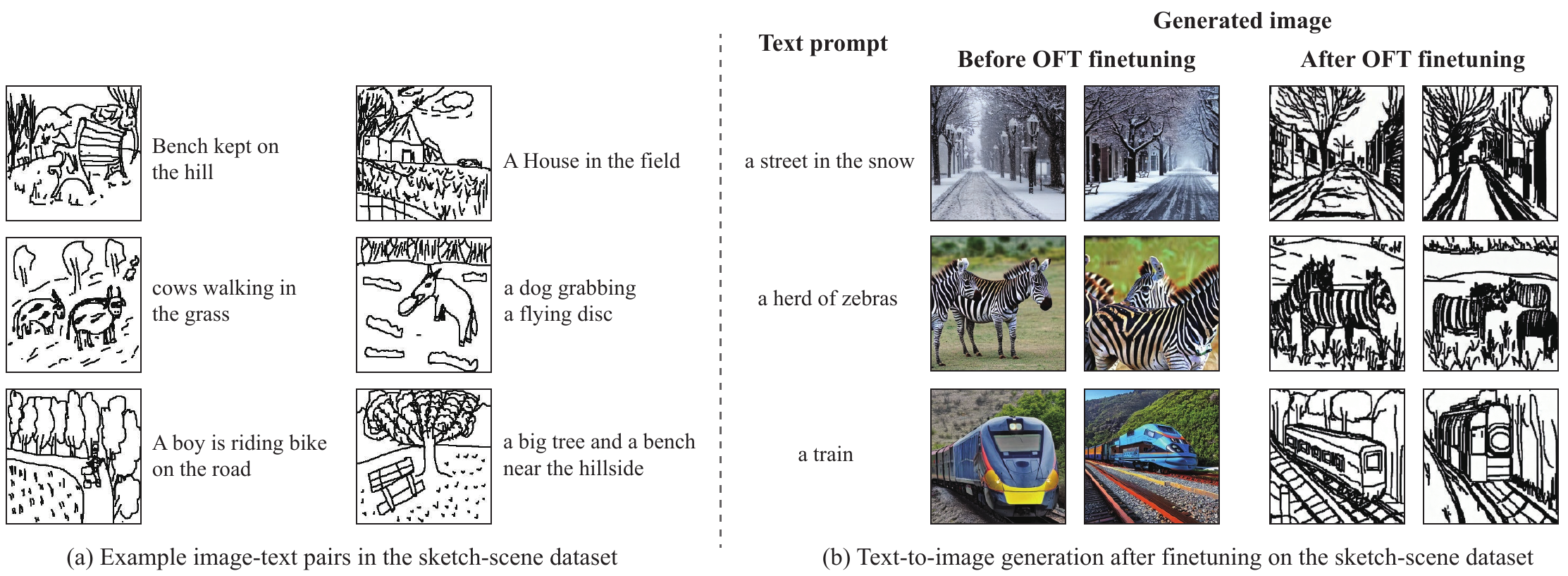}
    \vspace{-5mm}
    \caption{\scriptsize Direct OFT Finetuning of Stable Diffusion on the sketch-scene dataset.}
    \label{fig:sketch_scene}
\end{figure}

\begin{figure}[h]
    \centering
    \setlength{\abovecaptionskip}{10pt}
    \setlength{\belowcaptionskip}{0pt}
    \vspace{-2.5pt}
    \includegraphics[width=\textwidth]{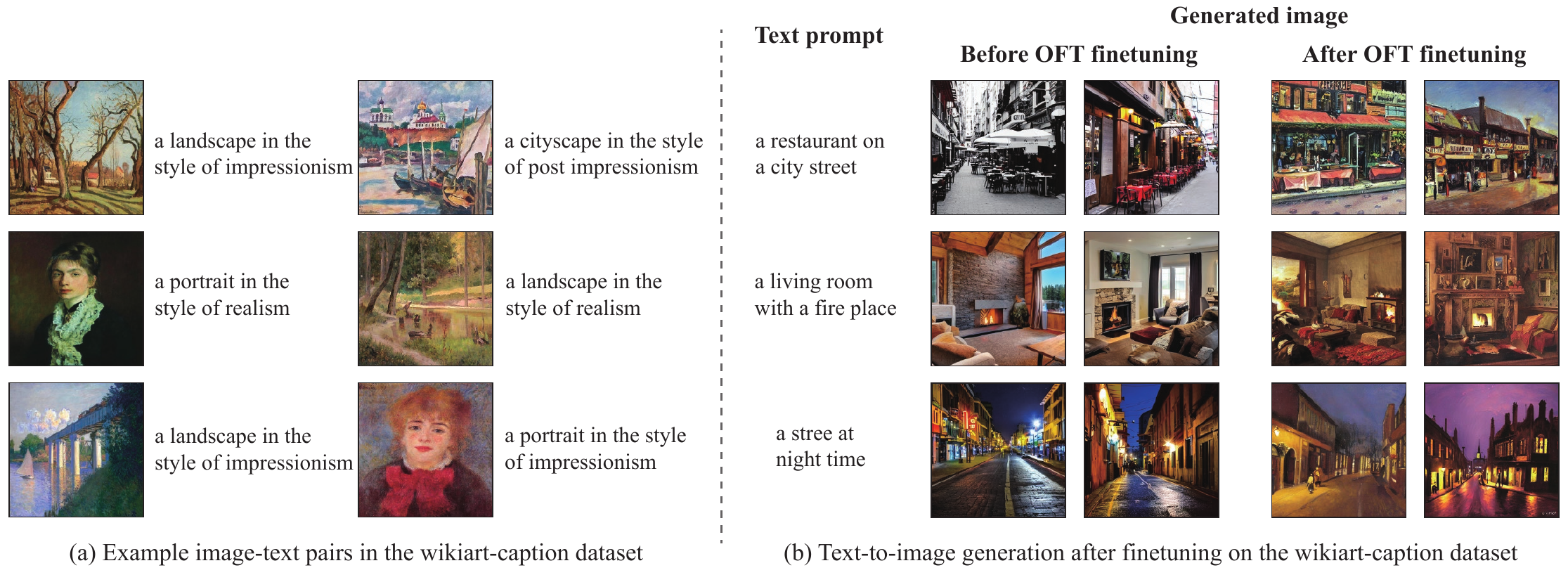}
    \vspace{-5mm}
    \caption{\scriptsize Direct OFT Finetuning of Stable Diffusion on the wikiart-caption dataset.}
    \label{fig:wikiart}
\end{figure}

Stable Diffusion can generate images based on the input text prompts. Without any adaptation, inputting text prompts to a pre-trained Stable Diffusion model will result in images that resemble natural images. We can finetune the pre-trained Stable Diffusion model on a custom dataset, to adapt the style of the generated images to the custom dataset. To demonstrate the effectiveness of orthogonal finetuning, we show qualitative results of adapting Stable Diffusion on the Sketch Scene dataset\footnote{\href{https://huggingface.co/datasets/zoheb/sketch-scene}{https://huggingface.co/datasets/zoheb/sketch-scene}} after finetuning for 20000 iterations in Figure~\ref{fig:sketch_scene} and on the Wikiart dataset\footnote{\href{https://huggingface.co/datasets/fusing/wikiart_captions}{https://huggingface.co/datasets/fusing/wikiart\_captions}} after finetuning for 30000 iterations in Figure~\ref{fig:wikiart}. We train on a single NVIDIA A100-SXM4-80GB GPU using a learning rate of $1 \times 10^{-4}$, batch size of $1$, and $4$ as the number of gradient accumulation steps.

\clearpage
\newpage
\section{Failure Cases}\label{app:failure}

\begin{figure}[h]
    \centering
    \setlength{\abovecaptionskip}{4pt}
    \setlength{\belowcaptionskip}{-10pt}
    \renewcommand{\captionlabelfont}{\scriptsize}
    \vspace{-2pt}
    \includegraphics[width=.99\textwidth]{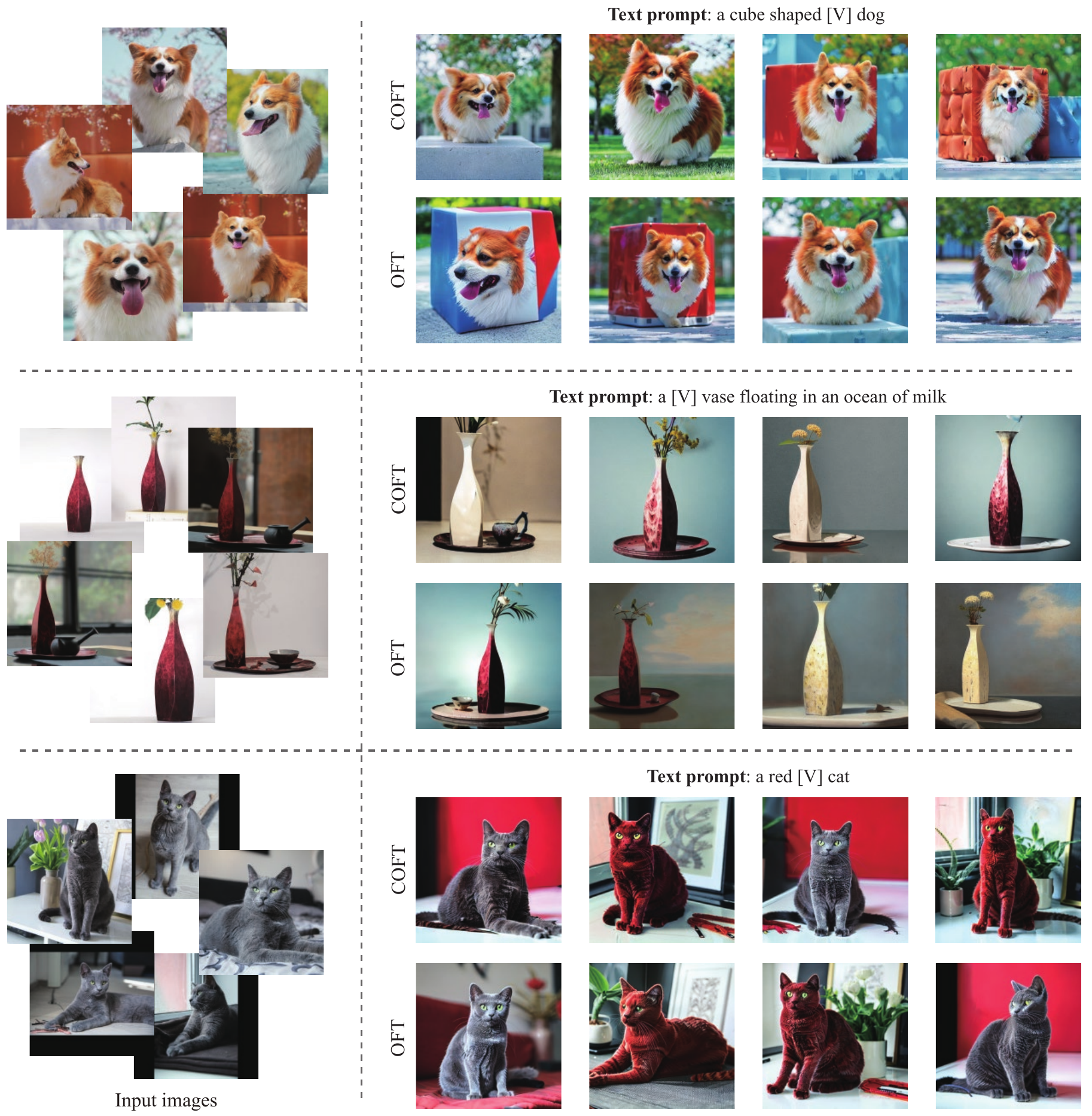}
    \caption{\scriptsize Some failure cases in subject-driven generation.\looseness=-1}
    \label{fig:failure_db}
\end{figure}

\begin{figure}[h]
    \centering
    \setlength{\abovecaptionskip}{4pt}
    \setlength{\belowcaptionskip}{-10pt}
    \renewcommand{\captionlabelfont}{\scriptsize}
    \vspace{-2pt}
    \includegraphics[width=.8\textwidth]{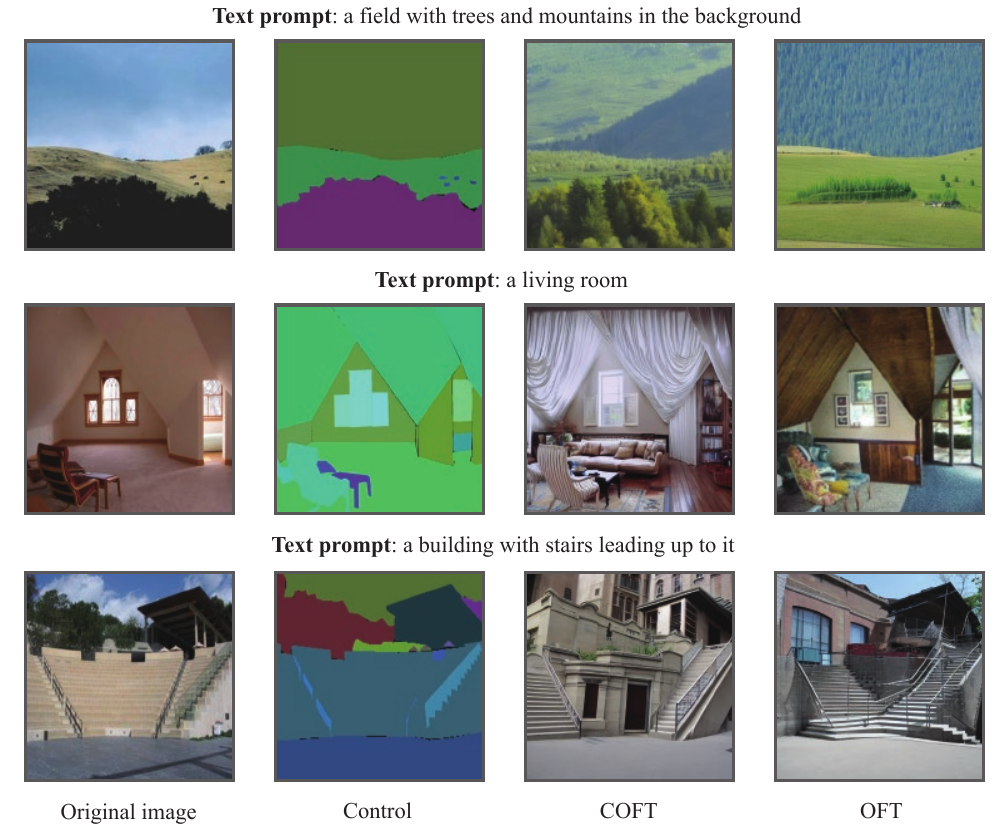}
    \caption{\scriptsize Some failure cases in controllable generation.\looseness=-1}
    \label{fig:failure_control}
\end{figure}

We also show a few failure cases of OFT and COFT. Figure~\ref{fig:failure_db} gives three failure cases in subject-driven generation. Figure~\ref{fig:failure_control} gives three failure cases in controllable generation. 

\subsection{Failure Cases in Subject-driven Generation}

In subject-driven generation, OFT and COFT will sometimes fail to ground the text attribute to the intended object. In the cat example, both OFT and COFT will sometimes generate other red objects, instead of generating a red cat.

\newpage
\subsection{Failure Cases in Controllable Generation}

Both OFT and COFT will sometimes hallucinate complicated structural details in a large region with the same semantics. Despite still being visually plausible, these generated images cannot match the original segmentation maps.

